\documentclass{article} 
\usepackage{iclr2021_conference,times}

\iclrfinalcopy

\usepackage[utf8]{inputenc} 
\usepackage[T1]{fontenc}    
\usepackage{hyperref}       
\usepackage{url}            
\usepackage{booktabs}       
\usepackage{amsfonts}       
\usepackage{nicefrac}       
\usepackage{microtype}      
\usepackage{multirow}
\usepackage{graphicx}
\usepackage{subfigure}
\usepackage{rotating}
\usepackage{mathtools}
\usepackage{overpic}
\usepackage[toc,page]{appendix}

\usepackage{color}
\definecolor{green}{rgb}{0, 0.5, 0}
\definecolor{orange}{rgb}{0.8, 0.6, 0.2}
\definecolor{red}{rgb}{1.0, 0.0, 0.0}
\definecolor{teal}{rgb}{0.0, 0.4, 0.4}
\definecolor{purple}{rgb}{0.65,0,0.65}
\definecolor{saffron}{rgb}{0.95,0.75,0.2}
\definecolor{turquoise}{rgb}{0.0,0.5,0.5}

\newcommand{\supl}[1]{{\emph{\color{black}#1}}}

\newcommand{\cred}[1]{{\color{red}{#1}}}
\newcommand{\bcred}[1]{{\color{red}{\textbf{#1}}}}

\newcommand{\cblue}[1]{{\color{blue}{#1}}}
\newcommand{\bcblue}[1]{{\color{blue}{\textbf{#1}}}}

\title{AReLU: Attention-based Rectified Linear Unit}

\author{Dengsheng Chen, \quad Jun Li, \quad Kai Xu\thanks{Corresponding author: Kai Xu (kevin.kai.xu@gmail.com)} \\
National University of Defense Technology \\
}

\begin{document}

\maketitle

\vspace{-12pt}
\begin{abstract}
Element-wise activation functions play a critical role in deep neural networks via affecting the expressivity power and the learning dynamics. Learning-based activation functions have recently gained increasing attention and success. We propose a new perspective of learnable activation function through formulating them with \emph{element-wise attention mechanism}. In each network layer, we devise an attention module which learns an element-wise, sign-based attention map for the pre-activation feature map. The attention map scales an element based on its sign. Adding the attention module with a rectified linear unit (ReLU) results in an amplification of positive elements and a suppression of negative ones, both with learned, data-adaptive parameters. We coin the resulting activation function Attention-based Rectified Linear Unit (AReLU). The attention module essentially learns an element-wise residue of the activated part of the input, as ReLU can be viewed as an identity transformation. This makes the network training more resistant to gradient vanishing. The learned attentive activation leads to well-focused activation of relevant regions of a feature map. Through extensive evaluations, we show that AReLU significantly boosts the performance of most mainstream network architectures with only two extra learnable parameters per layer introduced. Notably, AReLU facilitates fast network training under small learning rates, which makes it especially suited in the case of transfer learning and meta learning.
\end{abstract}\vspace{-8pt}
\vspace{-6pt}

\section{Introduction}
Activation functions, introducing nonlinearities to artificial neural networks, is essential to networks' expressivity power and learning dynamics.
Designing activation functions that facilitate fast training of accurate deep neural networks is an active area of research \citep{maas2013rectifier,goodfellow2013maxout,xu2015empirical,clevert2015fast,hendrycks2016gaussian,klambauer2017self,barron2017continuously,ramachandran2017searching}.
Aside from the large body of hand-designed functions, learning-based approaches recently gain more attention and success \citep{agostinelli2014learning,he2015delving,manessi2018learning,molina2019pad,1906.09529}.
The existing learnable activation functions are motivated either by relaxing/parameterizing a non-learnable activation function (e.g. Rectified Linear Units (ReLU) \citep{nair2010rectified}) with learnable parameters \citep{he2015delving}, or by seeking for a data-driven combination of a pool of pre-defined activation functions \citep{manessi2018learning}.
Existing learning-based methods make activation functions data-adaptive through introducing degrees of freedom and/or enlarging the hypothesis space explored.

In this work, we propose a new perspective of learnable activation functions through formulating them with \emph{element-wise attention mechanism}. A straightforward motivation of this is a straightforward observation that both activation functions and element-wise attention functions are applied as a network module of \emph{element-wise multiplication}. More intriguingly, learning element-wise activation functions in a neural network can intuitively be viewed as task-oriented attention mechanism~\citep{chorowski2015attention,xu2015show}, i.e., \emph{learning where (which element in the input feature map) to attend (activate) given an end task to fulfill}.
This motivates an arguably more interpretable formulation of \emph{attentive activation functions}.

Attention mechanism has been a cornerstone in deep learning. It directs the network to learn which part of the input is more relevant or contributes more to the output. There have been many variants of attention modules with plentiful successful applications. In natural language processing, vector-wise attention is developed to model the long-range dependencies in a sequence of word vectors~\citep{luong2015effective,vaswani2017attention}. Many computer vision tasks utilize pixel-wise or channel-wise attention modules for more expressive and invariant representation learning~\citep{xu2015show,chen2017sca}. Element-wise attention~\citep{bochkovskiy2020yolov4} is the most fine-grained where each element of a feature volume can receive different amount of attention.
Consequently, it attains high expressivity with neuron-level degrees of freedom.

Inspired by that, we devise for each layer of a network an element-wise attention module which learns a sign-based attention map for the pre-activation feature map. The attention map scales an element based on its sign. Through adding the attention and a ReLU module, we obtain Attention-based Rectified Linear Unit (AReLU) which amplifies positive elements and suppresses negative ones, both with learned, data-adaptive parameters. The attention module essentially learns an element-wise residue for the activated elements with respect to the ReLU since the latter can be viewed as an identity transformation. This helps ameliorate the gradient vanishing issue effectively. Through extensive experiments on several public benchmarks, we show that AReLU significantly boosts the performance of most mainstream network architectures with only two extra learnable parameters per layer introduced. Moreover, AReLU enables fast learning under small learning rates, making it especially suited for transfer learning. We also demonstrate with feature map visualization that the learned attentive activation achieves well-focused, task-oriented activation of relevant regions.

\section{Related Work}
\label{related}

\vspace{-8pt}
\paragraph{Non-learnable activation functions}
Sigmoid is a non-linear, saturated activation function used mostly in the output layers of a deep learning model. However, it suffers from the exploding/vanishing gradient problem. As a remedy, the rectified linear unit (ReLU)~\citep{nair2010rectified} has been the most widely used activation function for deep learning models with the state-of-the-art performance in many applications.
Many variants of ReLU have been proposed to further improve its performance on different tasks LReLU \citep{maas2013rectifier}, ReLU6 \citep{krizhevsky2010convolutional}, RReLU \citep{xu2015empirical}. Besides that, some specified activation functions also have been designed for different usages, such as CELU \citep{barron2017continuously}, ELU \citep{clevert2015fast}, GELU \citep{hendrycks2016gaussian}, Maxout \citep{goodfellow2013maxout}, SELU \citep{klambauer2017self}, (Softplus) \citep{glorot2011deep}, Swish \citep{ramachandran2017searching}.

\vspace{-8pt}
\paragraph{Learnable activation functions}
Recently, learnable activation functions have drawn more attentions. PReLU~\citep{he2015delving}, as a variants of ReLU, improves model fitting with little extra computational cost and overfitting risk.
Recently, PAU~\citep{molina2019pad} is proposed to not only approximate common activation functions but also learn new ones while providing compact representations with few learnable parameters. Several other learnable activation functions such as APL~\citep{agostinelli2014learning}, Comb~\citep{manessi2018learning}, SLAF~\citep{1906.09529} also achieve promising performance under different tasks.

\vspace{-8pt}
\paragraph{Attention Mechanism}
Vector-Wise Attention Mechanism (VWAM) has been widely applied in Natural Language Processing (NLP) tasks~\citep{sat,luong2015effective,1409.0473,vaswani2017attention,1711.02132}. VWAM learns which vector among a sequence of word vectors is the most relevant to the task in hand.
Channel-Wise Attention Mechanism (CWAM) can be regarded as an extension of VWAM from NLP to Vision tasks~\citep{tang2019multichannel,tang2020multi,1907.10830}. It learns to assign each channel an attentional value.
Pixel-Wise Attention Mechanism (PWAM) is also widely used in vision~\citep{tang2019attention,tang2019attentiongan}.
Element-Wise Attention Mechanism (EWAM) assigns different values to each element without any spatial/channel constraint. The recently proposed YOLOv4~\citep{bochkovskiy2020yolov4} is the first work that introduces EWAM implemented by a convolutional layer and sigmoid function. It achieves the state-of-the-art performance on object detection.
We introduce a new kind of EWAM for learnable activation function.


\section{Method}
We start by describing attention mechanism and then
introduce element-wise sign-based attention mechanism based on which AReLU is defined.
The optimization of AReLU then follows.

\subsection{Attention Mechanism}
Let us denote $V = \left\{v_i\right\} \in \mathbb{R}^{D_v^1 \times D_v^2 \times \cdots}$ a tensor representing input a data or feature volume.
Function $\Phi$, parameterized by $\Theta = \left\{\theta_i\right\}$, is used to compute an attention map $S = \left\{s_i\right\} \in \mathbb{R}^{D_v^{\theta(1)} \times D_v^{\theta(2)} \times \cdots}$ over a subspace of $V$
(let $\theta(\cdot)$ denote a correspondence function for the indices of dimension):
$$
s_i = \Phi(v_i, \Theta).
$$
$\Phi$ can be implemented by a neural network with $\Theta$ being its learnable parameters.

\begin{figure}[t]
	\centering
	\includegraphics[width=\textwidth]{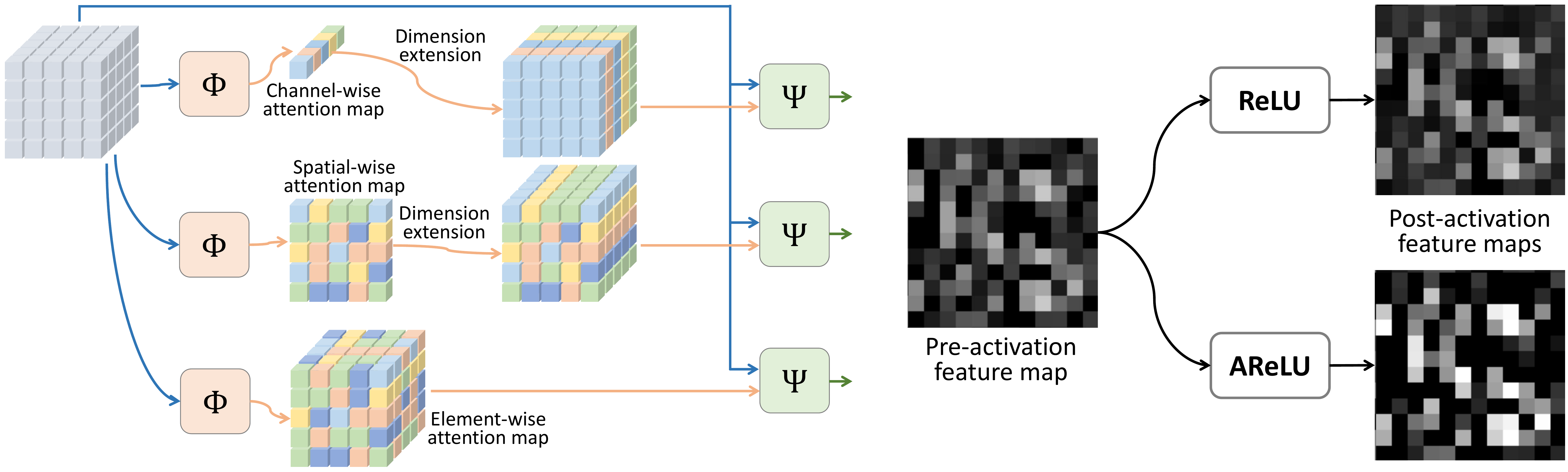}
	\caption{Left: An illustration of attention mechanisms with attention map at different granularities. Right: An visualization of pre-activation and post-activation feature maps obtained with ReLU and AReLU on a testing image of the handwritten digit dataset MNIST \cite{lecun1998gradient}.}
	\label{fig:attention}\vspace{-12pt}
\end{figure}

We can modulate the input $V$ with the attention map $S$ using a function $\Psi$, obtaining the output $U = \left\{u_i \right\} \in \mathbb{R}^{D_v^1 \times D_v^2 \times \cdots}$:
$$
u_i = \Psi(v_i, s_i).
$$
$\Psi$ is an element-wise multiplication. In order to perform element-wise multiplication, one needs to first extend $S$ to the full dimension of $V$.
We next review various attention mechanisms with attention map at different granularities.
Figure~\ref{fig:attention}(left) gives an illustration of various attention mechanisms.

\textbf{Vector-wise Attention Mechanism}
In NLP, attention maps are usually computed over different word vectors.
In this case, $V = \left\{v_i\right\} \in \mathbb{R}^{N \times D}$ represents a sequence of $N$ feature vectors with dimension $D$. $S = \left\{s_i\right\} \in \mathbb{R}^{N} $ is a sequence of attention values for the corresponding vectors.

\textbf{Channel-wise Attention Mechanism}
In computer vision, a feature volume $V = \left\{v_i \right\} \in \mathbb{R}^{W \times H \times C}$ has a spatial dimension of $W \times H$ and a channel dimension of $C$. $S = \left\{s_i \right\} \in \mathbb{R}^{C}$ is an attention map over the $C$ channels. All elements in each channel share the same attention value.

\textbf{Spatial-wise Attention Mechanism}
Considering again $V = \left\{v_i \right\} \in \mathbb{R}^{W \times H \times C}$ with a spatial dimension of $W \times H$. $S = \left\{s_i \right\} \in \mathbb{R}^{W \times H}$ is an attention map over the spatial dimension. All channels of a given spatial location share the same attention value.

\textbf{Element-wise Attention Mechanism}
Given a feature volume $V = \left\{v_i \right\} \in \mathbb{R}^{W \times H \times C}$ containing $W \times H \times C$ elements, we compute an attention map over the whole volume (all elements), i.e., $S = \left\{s_i \right\} \in \mathbb{R}^{W \times H \times C}$, so that each element has an independent attention value.

\subsection{ELement-wise Sign-based Attention (ELSA)}
We propose, ELSA, a new kind of element-wise attention mechanism which is used to define our attention-based activation.
Considering a feature volume $V = \left\{v_i \right\} \in \mathbb{R}^{W \times H \times C}$, we compute an element-wise attention map $S = \left\{s_i \right\} \in \mathbb{R}^{W \times H \times C}$:
$$
s_i = \Phi(v_i, \Theta) = \left\{\begin{matrix}
C(\alpha), & v_i < 0\\
\sigma(\beta), & v_i \geq 0
\end{matrix}\right.
$$
where $\Theta = \left\{\alpha, \beta\right\} \in \mathbb{R}^{2}$ is learnable parameters. $C(\cdot)$ clamps the input variable into $\left[0.01, 0.99\right]$. $\sigma$ is the sigmoid function. The modulation function $\Psi$ is defined as:
$$
u_i = \Psi(v_i, s_i) = s_i v_i.
$$
In ELSA, positive and negative elements receive different amount of attention determined by the two parameters $\alpha$ and $\beta$, respectively. Therefore, it can also be regarded as sign-wise attention mechanism.
With only two learnable parameters, ELSA is light-weight and easy to learn.



\subsection{AReLU: Attention-based Rectified Linear Units}
We represent the function $\Phi$ in ELSA with a network layer with learnable parameters $\alpha$ and $\beta$:
$$
\mathcal{L}(x_i, \alpha, \beta) = \left\{
\begin{matrix}
	C(\alpha) x_i,        & x_i < 0    \\
	\sigma(\beta) x_i, & x_i \geq 0
\end{matrix}
\right.
$$
where $X = \left\{x_i \right\}$ is the input of the current layer.
In constructing an activation function with ELSA, we combine it with the standard Rectified Linear Units
$$
\mathcal{R}(x_i)=\left\{\begin{matrix}
0, & x_i < 0\\
x_i, & x_i \geq 0
\end{matrix}\right.
$$
Adding them together leads to a learnable activation function:
$$
\mathcal{F}(x_i, \alpha, \beta)= \mathcal{R}(x_i) + \mathcal{L}(x_i, \alpha, \beta) =\left\{\begin{matrix}
C(\alpha) x_i, & x_i < 0\\
(1 + \sigma(\beta)) x_i, & x_i \geq 0
\end{matrix}\right.
$$
This combination amplifies positive elements and suppresses negative ones based on the learned scaling parameters $\beta$ and $\alpha$, respectively.
Thus, ELSA learns an element-wise residue for the activated elements w.r.t. ReLU which is an identity transformation, which helps ameliorate gradient vanishing.

\subsection{The Optimization of AReLU}
AReLU can be trained using back-propagation jointly with all other network layers.
The update formulation of $\alpha$ and $\beta$ can be derived with the chain rule.
Specifically, the gradient of $\alpha$ is:
$$
\frac{\partial \mathcal{E} }{\partial \alpha} = \frac{\partial \mathcal{E}}{\partial \mathcal{F}(x_i, \alpha, \beta)} \frac{\partial \mathcal{F}(x_i, \alpha, \beta))}{\partial \alpha}
$$
where $\mathcal{E}$ is the error function to be minimized.
The term $\frac{\partial \mathcal{E}}{\partial \mathcal{F}(x_i, \alpha, \beta)}$ is the gradient propagated from the deeper layer. The gradient of the activation of $X$ with respect to $\alpha$ is given by:
$$
\frac{\partial \mathcal{F}(X, \alpha, \beta)}{\partial \alpha} = \sum_{x_i < 0} x_i
$$
Here, the derivative of the clamp function $C(\cdot)$ is handled simply by detaching the gradient back-propagation when $\alpha<0.01$ or $\alpha>0.99$.

The gradient of the activation of $X$ with respect to $\beta$ is:
$$
\frac{\partial \mathcal{F}(X, \alpha, \beta)}{\partial \beta} = \sum_{x_i \geq 0} \sigma(\beta)(1-\sigma(\beta)) x_i
$$

The gradient of the activation with respect to input $x_i$ by:
$$
\frac{\partial \mathcal{F}(x_i, \alpha, \beta)}{\partial x_i}=\left\{\begin{matrix}
\alpha, & x_i < 0\\
1 + \sigma(\beta), & x_i \geq 0
\end{matrix}\right.
$$


It can be found that AReLU amplifies the gradients propagated from the downstream when the input is activated since $1 + \sigma(\beta) > 1$; it suppresses the gradients otherwise.
On the contrary, there is no such amplification effect in the standard ReLU and its variants (e.g., PReLu \citep{he2015delving}) --- only suppression is available.
The ability to amplify the gradients over the activated input helps avoiding gradient vanishing,
and thus speeds up the training convergence of the model (see Figure~\ref{fig:mnistplot}).
Moreover, the amplification factor is learned to dynamically adapt to the input and is confined with the sigmoid function. This makes the activation more data-adaptive and stable (see Figure~\ref{fig:attention}(right) for a visual comparison of post-activation feature maps by AReLU and ReLU).
The suppression part is similar to PReLu which learns the suppression factor for ameliorating zero gradients.

AReLU introduces a very small number of extra parameters which is $2L$ for an $L$-layer network.
The computational complexity due to AReLU is negligible for both forward and backward propagation.

Note that the gradients of $\alpha$ and $\beta$ depend on the entire feature volume $X$.
This means that ELSA can be regarded as a global attention mechanism: Although the attention map is computed in an element-wise manner, the parameters are learned globally accounting for the impact of the full feature volume.
This makes our AReLU more data-adaptive and hence the whole network more expressive.

We adopt the momentum method for updating $\alpha$ and $\beta$:
$$
\Delta \alpha := \mu \Delta \alpha + \epsilon \frac{\partial \mathcal{E} }{\partial \alpha}, \quad
\Delta \beta := \mu \Delta \beta + \epsilon \frac{\partial \mathcal{E} }{\partial \beta},
$$
where $\mu$ is the momentum and $\epsilon$ the learning rate.
It is worth noticing that a weight decay ($L_2$ regularization) tends to push $\alpha$ to zero.
Confining $\alpha$ within $\left[0.01, 0.99\right]$ mitigates this issue.


\section{Experiments}
We first study the robustness of AReLU in terms of parameter initialization.
We then evaluate convergence of network training with different activation functions on two standard classification benchmarks (MNIST~\citep{lecun1998gradient}) and CIFAR100~\citep{krizhevsky2009learning}.
We compare AReLU with 18 different activation functions including 13 non-learnable ones and 5 learnable ones; see the list in Table~\ref{tab:mnist-1st-mean}.
The number of learnable parameters for each learnable activation function are also given in the table.
In the end, we also demonstrate the advantages of AReLU in transfer learning.
\supl{Please refer to supplemental material for more results and experiments details.}

\subsection{Initialization of Learnable Parameter $\alpha$ and $\beta$}

\begin{figure}[t]
	\centering
	\begin{overpic}[width=\textwidth]{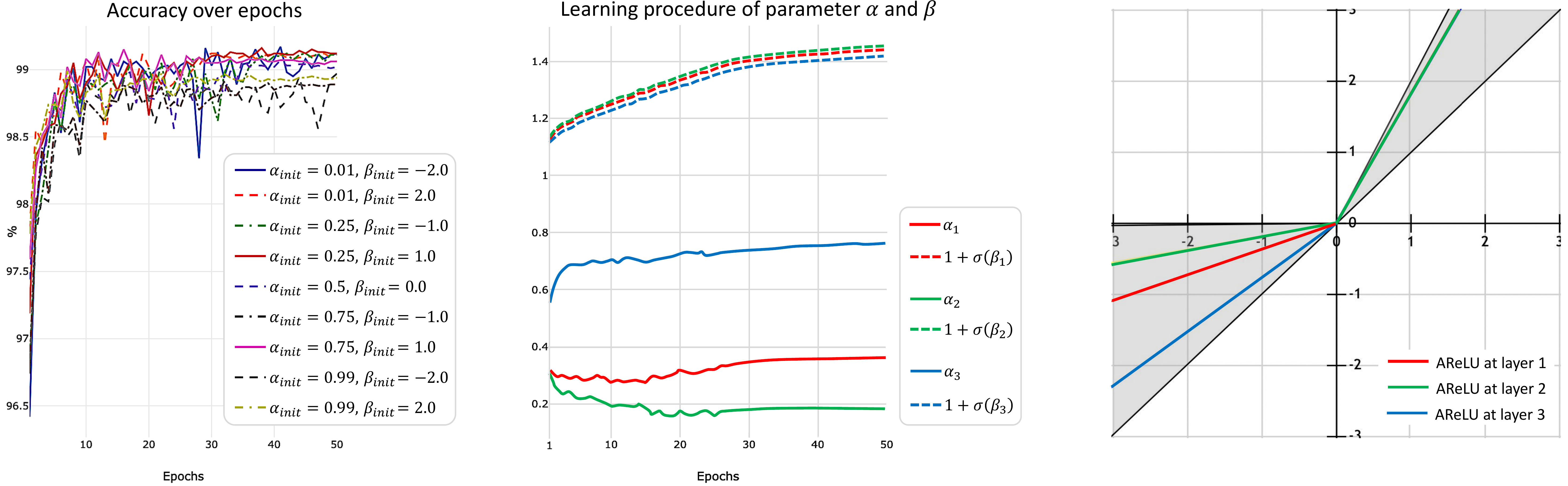}
    \put(15,32){\small (a)}
    \put(47,32){\small (b)}
    \put(84,32){\small (c)}
    \end{overpic}\vspace{-8pt}
	\caption{(a): Plot of accuracy over epochs for networks trained with different initialization of $\alpha$ and $\beta$. A larger initial $\beta$ leads to faster convergence and higher accuracy is obtained when $\alpha$ is initialized to $0.25$ or $0.75$. (b): The learning procedure of $\alpha$ and $\beta$ which are initialized to $0.25$ and $1.0$, respectively. (c): The learned final AReLU's for the three convolutional layers of the MNIST-Conv network. The shaded region gives the range of AReLU curves.}
	\label{alpha_beta}\vspace{-8pt}
\end{figure}

For evaluation purpose, we design a neural network (MNIST-Conv) with three convolutional layers each followed by a max-pooling layer and an AReLU, and finally a fully connected layer followed by a softmax layer.
\supl{Details of this network can be found in the supplemental material.}
The experiment on parameter initialization is conducted with MNIST-Conv over the MNIST dataset.
As shown in Figure~\ref{alpha_beta}(a), AReLU is insensitive to the initialization of $\alpha$ and $\beta$.
Different initial values result in close convergence rate and classification accuracy.
Generally, a large initial value of $\beta$ can speed up the convergence.
Figure~\ref{alpha_beta}(b) shows the learning procedure of the two parameters and (c) plots the learned final AReLU's for the three convolutional layers.
%
%
In the following experiments, we initialize $\alpha=0.9$ and $\beta=2.0$ by default.


\subsection{Convergence on MNIST}

\begin{figure}
	\centering
	\includegraphics[width=0.9\textwidth]{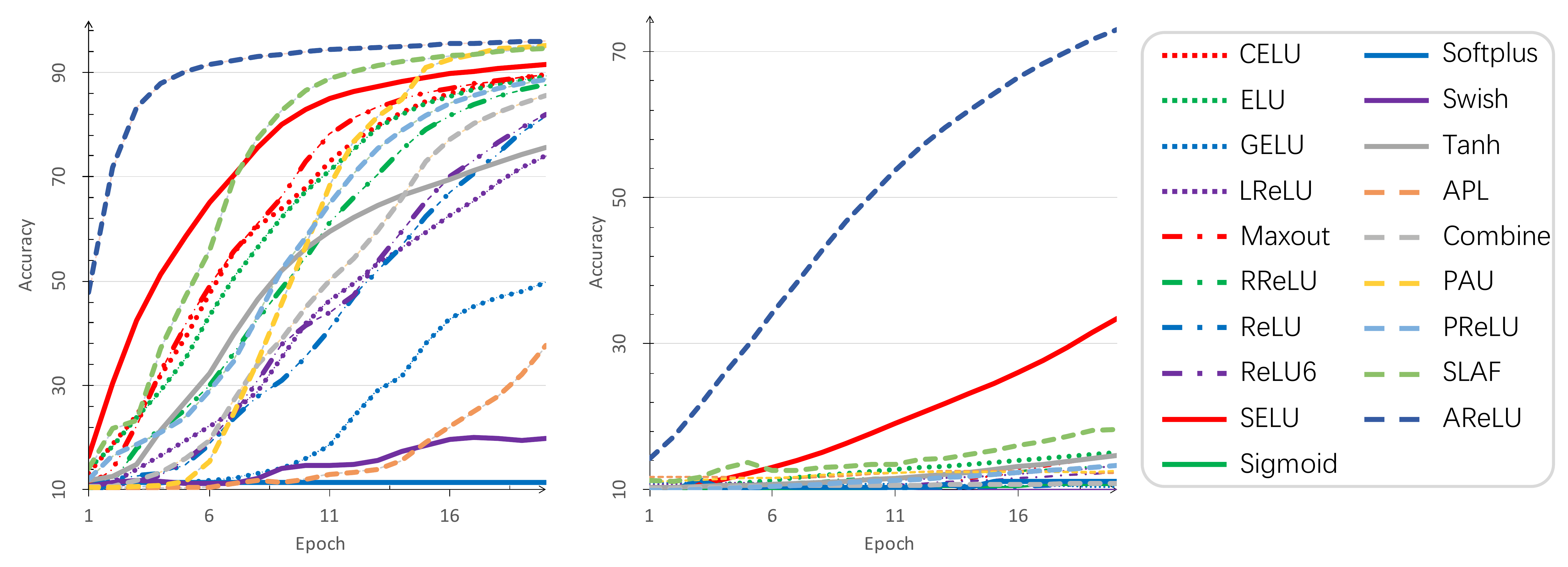}\vspace{-12pt}
	\caption{Plots of mean testing accuracy (\%) on MNIST for five-time trainings of MNIST-Conv over increasing training epochs. The training is conducted using SGD with small learning rates (left: $1\times 10^{-4}$, right: $1\times 10^{-5}$).}
	\label{fig:mnistplot}\vspace{-8pt}
\end{figure}

On the MNIST dataset, we evaluate MNIST-Conv implemented with different activation functions and trained with the ADAM or SGD optimizer.
The activation function is placed after each max-pooling layers.
We compare AReLU with both learnable and non-learnable activation functions under different learning rates of $1\times 10^{-2}$, $1\times 10^{-3}$, $1\times 10^{-4}$, and $1\times 10^{-5}$.
To compare the convergence speed of different activation functions, we report the accuracy after the first epoch, again taking the mean over five times training; see Table~\ref{tab:mnist-1st-mean}.
In the table, we report the improvement of AReLU over the best among other non-learnable and learnable methods.
In Figure~\ref{fig:mnistplot}, we plot the mean accuracy over increasing number of training epochs.


As shown in Table~\ref{tab:mnist-1st-mean}, AReLU outperforms most existing non-learnable and learnable activation functions in terms of convergence speed and final classification accuracy on MNIST.
A note-worthy phenomenon is that AReLU can achieve a more effective training with a small learning rate (see the significant improvement when the learning rate is $1\times 10^{-4}$ or $1\times 10^{-5}$) than the alternatives. This can also be observed from Figure~\ref{fig:mnistplot}.
Generally, smaller learning rates would cause lower learning efficiency since the vanishing gradient issue is intensified in such case.
AReLU can overcome this difficulty thanks to its gradient amplification effect.
Efficient learning with a small learning rate is very useful in transfer learning where a pre-trained model is usually fine-tuned on a new domain/dataset with a small learning rate which is difficult for most existing deep networks. Section~\ref{sec:transfer} will demonstrate this application of AReLU.

\begin{table}
	\centering\small
	\caption{Mean testing accuracy (\%) on MNIST for five trainings of MNIST-Conv after the \emph{first epoch} with different optimizers and learning rates. We compare AReLU with $13$ non-learnable and $5$ learnable activation functions. The number of parameters per activation unit are listed beside the name of the learnable activation functions. The best numbers are shown in bold text with blue color for non-learnable methods (the upper part of the table) and red for learnable ones (the lower part). At the bottom of the table, we report the improvement of AReLU over the best among other non-learnable and learnable methods, in blue and red color respectively.}\vspace{8pt}
	\label{tab:mnist-1st-mean}
	\begin{tabular}{l|rr|rr|rr|rr}
		Learning Rate & \multicolumn{2}{c|}{$1\times 10^{-2}$}                          & \multicolumn{2}{c|}{$1\times 10^{-3}$}                        & \multicolumn{2}{c|}{$1 \times 10 ^{-4}$}                       & \multicolumn{2}{c}{$1\times 10^{-5}$}                      \\ \hline
		Optimizer                               & Adam                    & SGD                     & Adam                    & SGD                     & Adam                    & SGD                     & Adam                    & SGD                     \\ \hline\hline
		CELU \citeyearpar{barron2017continuously}       & 97.76                   & 96.12                   & 96.21                   & 62.81                   & 84.01                   & 13.07                   & 24.84                   & 9.60                    \\
		ELU \citeyearpar{clevert2015fast}               & 97.82                   & 96.17                   & 96.22                   & 58.10                   & \bcblue{85.67}                   & 14.07          & 19.77                   & 10.13                   \\
		GELU \citeyearpar{hendrycks2016gaussian}                   & \bcblue{98.49}          & 94.90                   & 95.79                   & 12.55                   & 83.72                   & 11.49                   & 15.20                   & \bcblue{10.92}                   \\
		LReLU \citeyearpar{maas2013rectifier}       & 97.80                   & 95.59                   & 95.86                   & 35.90                   & 84.08                   & 10.28                   & 15.41                   & 10.73                   \\
		Maxout \citeyearpar{goodfellow2013maxout}       & 97.04                   & 95.81                   & 96.14                   & 71.75                   & 84.81                   & 10.79                   & 18.83                   & 9.06                    \\
		ReLU \citeyearpar{nair2010rectified}            & 97.75                   & 95.02                   & 95.40                   & 36.01                   & 84.02                   & 10.68                   & 15.25                   & 8.73                    \\
		ReLU6 \citeyearpar{krizhevsky2010convolutional} & 97.77                   & 95.32                   & 96.09                   & 43.42                   & 81.39                   & 10.23                   & 14.33                   & 9.56                    \\
		RReLU \citeyearpar{xu2015empirical}             & 98.09                   & 95.88                   & 95.65                   & 53.33                   & 84.51                   & 9.57                    & 16.53                   & 10.28                   \\
        SELU \citeyearpar{klambauer2017self}            & 97.25                   & \bcblue{96.52}                   & \bcblue{96.61}          & \bcblue{82.36}          & 85.36            & \bcblue{16.49}   &   \bcblue{30.04} & 9.59                    \\
		Sigmoid                                 & 47.16                   & 11.04                   & 83.59                   & 11.35                   & 11.37                   & 9.92                    & 10.52                   & 10.10                   \\
		Softplus \citeyearpar{glorot2011deep}           & 96.38                   & 90.90                   & 93.83                   & 11.14                   & 51.83                   & 9.19                    & 10.21                   & 9.89                    \\
		Swish \citeyearpar{ramachandran2017searching}                  & 98.10                   & 94.02                   & 95.91                   & 11.44                   & 83.91                   & 10.69                   & 11.39                   & 9.47                    \\
		Tanh                                    & 96.93                   & 94.22                   & 96.45                   & 57.70                   & 79.25                   & 11.73                   & \textit{27.05}          & 10.31                   \\ \hline \hline
		APL \citeyearpar{agostinelli2014learning} ($2$)       & 97.00                   & 95.71                   & 94.67                   & 17.81                   & 76.73                   & 9.39                    & 13.28                   & 11.83 \\
		Comb \citeyearpar{manessi2018learning} ($1$)       & \bcred{98.28} & 95.97                   & 95.79                   & 35.95                   & 83.91                   & 10.59                   & 20.22                   & 10.18                   \\
		PAU \citeyearpar{molina2019pad} ($10$)                 & 98.17                   & \bcred{97.67}     & 96.73 & 40.11                   & 87.08 & 10.54                   & 14.49                   & 11.11                   \\
		PReLU \citeyearpar{he2015delving} ($1$)               & 98.22       & 95.72                   & 95.87                   & 45.73                   & 85.81          & 12.08                   & 14.51                   & 9.88                    \\
		SLAF \citeyearpar{1906.09529} ($2$)        & 96.30         & 97.07      & 95.32              & 83.35    & 72.67                   & 14.12                   & 10.04                   & 11.32          \\  \hline
		AReLU ($2$)                                   & 98.00                   & 97.30 & \bcred{97.13}          & \bcred{93.13}          & \bcred{90.44}          & \bcred{47.78}          & \bcred{38.39}          & \bcred{14.25}          \\
		\cblue{Improvement}      & \cblue{$-0.49$}     & \cblue{$+0.78$}          & \cblue{$+0.52$}          & \cblue{$+10.77$}     & \cblue{$+4.77$}      & \cblue{+$31.29$}       & \cblue{$+8.35$}      & \cblue{$+3.33$}\\
        \cred{Improvement}       & \cred{$-0.28$}      & \cred{$-0.37$}          & \cred{$+0.40$}          & \cred{$+9.78$}     & \cred{$+3.36$}     & \cred{$+33.66$}    & \cred{$+18.17$}      & \cred{$+2.42$}
	\end{tabular}\vspace{-16pt}
\end{table} 

	


\subsection{Convergence on CIFAR100}


In order to better demonstrate the effect of ELSA, we regard the ReLU, without ELSA, as our baseline. For plot clarity, we choose to compare only with those most representative competitive activation functions including PAU, SELU, ReLU, LReLU (LReLU), and PReLU.
\supl{More results can be found in the supplemental material.}
We evaluate the performance of AReLU with five different mainstream network architectures on CIFAR100.
We use the SGD optimizer and follow the training configuration in~\citep{pereyra2017regularizing}: The learning rate is $0.1$, the batch size is $64$, the weight decay is $5\times10^{-4}$, and the momentum is $0.9$.

\begin{figure}[t]
	\centering
	\includegraphics[width=\textwidth]{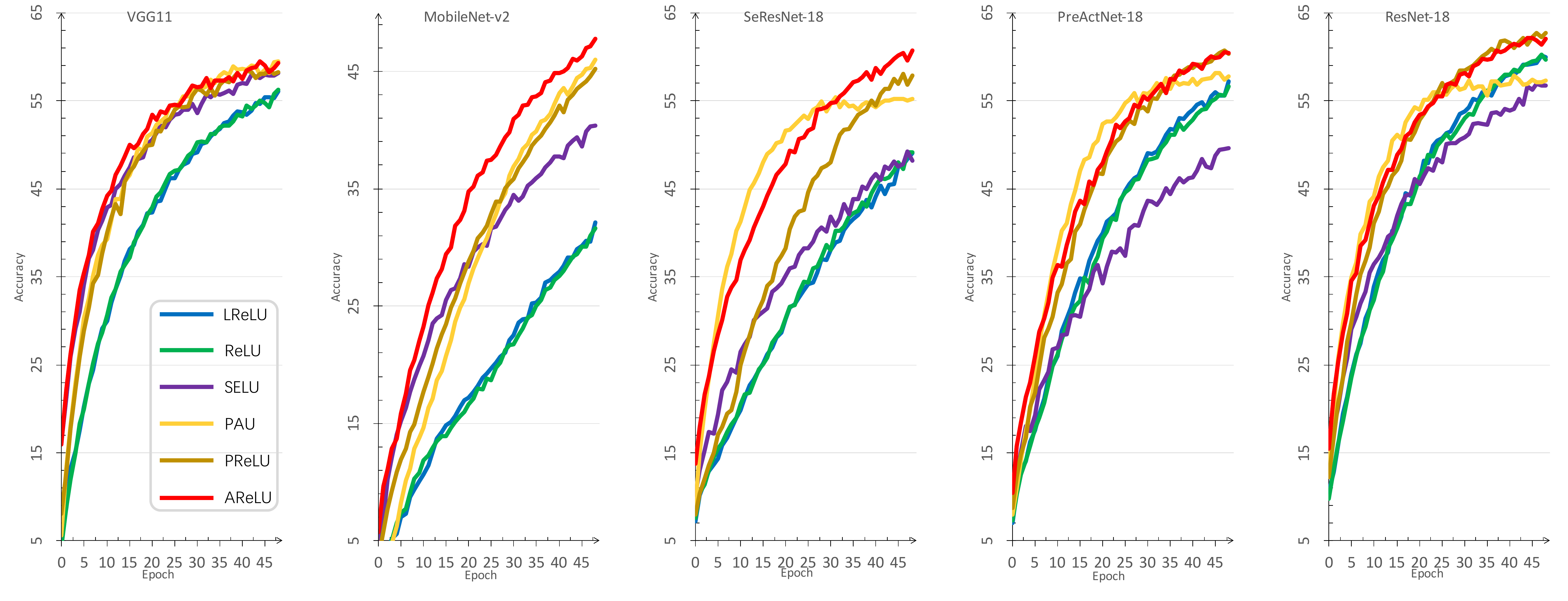}\vspace{-8pt}
	\caption{Plots of mean testing accuracy (\%) on CIFAR100 over increasing training epochs, using different network architectures. The training is conducted using SGD with a learning rate of $0.1$.}
	\label{fig:converge}\vspace{-8pt}
\end{figure} 

The results are plotted in Figure~\ref{fig:converge}.
Learnable activation functions generally have a faster convergence compared to non-learnable ones.
AReLU achieves a faster convergence speed for all the five network architectures. It is worth to note that though PAU can achieve a faster convergence at the beginning in some networks such as SeResNet-18, it tends to overfit later with a fast saturation of accuracy. AReLU avoids such overfitting with smaller number of parameters than PAU (2 vs 10).

\begin{figure}[t]
	\centering
	\includegraphics[width=0.9\textwidth]{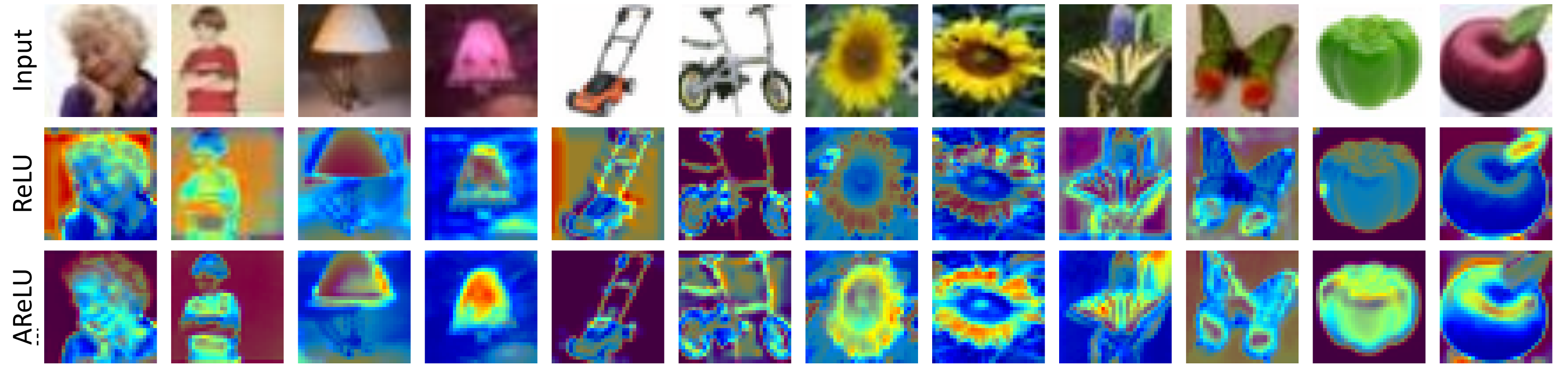}\vspace{-12pt}
	\caption{Grad-CAM visualization of feature maps extracted by ResNet-18 with AReLU and ReLU. The first row is the testing images of CIFAR100.}
	\label{gradcam}\vspace{-8pt}
\end{figure}

We also conduct a qualitative analysis of AReLU by visualizing the learned feature maps using Grad-CAM \cite{selvaraju2017grad} using testing images of CIFAR100.
Grad-CAM is a recently proposed network visualization method which utilizes gradients to depict the importance of the spatial locations in a feature map.
Since gradients are computed with respect to a specific image class,
Grad-CAM visualization can be regarded as a task-oriented attention map.
In Figure~\ref{gradcam}, we visualize the first-layer feature map of ResNet-18.
As shown in the figure, the feature maps learned with AReLU leads to semantically more meaningful activation of regions with respect to the target class. This is due to the data-adaptive, attentive ability of AReLU.

\subsection{Performance in Transfer Learning}
\label{sec:transfer}


\begin{table}[t]
	\centering\small
	\caption{Test accuracy (\%) on SVHN by MNIST-Conv models (implemented with different activation functions) trained directly on SVHN (no pretrain), trained on MNIST but not finetuned (no finetune), as well as pretrained on MNIST and finetuned on SVHN for 5, 10 and 20 epoches. The left part of the table is non-learnable activation functions and the right learnable ones.}\vspace{8pt}
	\scalebox{1}{
    \setlength{\tabcolsep}{0.75mm}{
	\begin{tabular}{l||c|c|c|c|c|c||c|c|c|c|c|c}
	Setting & ELU   & GELU  & Maxout & ReLU  & SELU  & Softplus & APL   & Comb  & PAU   & PReLU & SLAF  &AReLU  \\ \hline\hline
	no pretrain & 19.59 & 19.59    & 23.01  & 19.58 & 19.58 & 19.58   & 19.58    & 19.58 & 19.58 & 19.58 & 19.58   & \textbf{24.95} \\ \hline
    no finetune  & 31.95 & \textbf{37.38}  & 36.52  & 36.87 & 32.57 & 14.39    & 36.20 & 35.89   & 24.67 & 33.45 & 35.74 & 31.91          \\ \hline
	f.t. 5 epochs     & 70.08 & 69.19 & 70.18  & 69.76 & 72.81 & 65.81    & 71.73 & 69.63 & 75.24 & 66.13 & 75.91 & \textbf{76.68} \\ \hline
	f.t. 10 epochs    & 70.83 & 71.69 & 71.31  & 71.38 & 72.11 & 69.14    & 73.51 & 70.31 & 76.26 & 67.63 & 76.21 & \textbf{78.12} \\ \hline
	f.t. 20 epochs    & 71.48 & 70.34 & 73.29  & 72.06 & 71.55 & 71.01    & 72.41 & 72.55 & 74.46 & 71.99 & 74.38 & \textbf{78.48} \\ \hline
	\end{tabular}
	}}
	\label{tab:transfer}\vspace{-16pt}
\end{table}

We evaluate transfer learning of MNIST-Conv with different activation functions between two datasets: MNIST and SVHN\footnote{http://ufldl.stanford.edu/housenumbers/}.
The data preprocessing for adapting the two datasets follows~\citep{shin2017continual}.
We train three models and test them on SVHN: 1) one is trained directly on SVHN without any pretraining, 2) one trained on MNIST but not finetuned on SVHN, and 3) one pretrained on MNIST and finetuned on SVHN.
In pretraining, we train MNIST-Conv using SGD with a learning rate of $0.01$ for $20$ epochs which is sufficient for all model variants to converge.
In finetuning, we train the model on SVHN with a learning rate of $1\times10^{-5}$, using SGD optimizer for $100$ epochs.


The testing results on SVHN are reported in Table~\ref{tab:transfer} where we compare AReLU with several competitive alternatives.
Without pretraining, it is hard to obtain a good accuracy on the difficult task of SVHN. Nevertheless, MNIST-Conv with AReLU performs the best among all alternatives; some activation functions even failed in learning. In the setting of transfer learning (pretrain + finetune), AReLU outperforms all other activation functions for different amount of pretraining, thanks to it high learning efficiency with small learning rates.


\subsection{Performance in Meta Learning}
\label{sec:meta}
We evaluate the meta learning performance of MNIST-Conv with the various activation functions based on the MAML framework~\cite{finn2017model}.
MAML is a fairly general optimization-based algorithm compatible with any model that learns through gradient descent. It aims to obtain meta-learning parameters from similar tasks and adapt the parameters to novel tasks with the same distribution using a few gradient updates.
In MAML, model parameters are explicitly trained such that a small number of gradient updates over a small amount of training data from the novel task could lead to good generalization performance on that task.
We expect that the fast convergence of AReLU would help MAML to adapt a model to a novel task more efficiently and with better generalization.
We set the fast adaption steps as $5$ and use $32$ tasks for each steps. We train the model for $100$ iterations with a learning rate of $0.005$. We report in Table~\ref{tab:meta} the final test accuracy for different activation functions on a 5-ways-1-shots task and a 5-ways-5-shots task, respectively.
The results show that AReLU shows clear advantage compared to the alternative activation functions.
One noteworthy phenomenon is the performance of PAU~\citep{molina2019pad}: It performs well in other evaluations but not on meta learning which is probably due to its overfitting-prone nature.

\begin{table}[t]
    \centering\small
    \caption{Test accuracy (\%) on MNIST by MAML with MNIST-Conv models implemented with different activation functions. The performance is compared on a 5-ways-1-shots task and a 5-ways-5-shots task, respectively.}\vspace{2pt}
    \scalebox{1}{
    \setlength{\tabcolsep}{0.75mm}{
    \begin{tabular}{c||c|c|c|c|c|c||c|c|c|c|c|c}
    (ways, shots) &  ELU            & GELU  & Maxout & ReLU   & SELU    & Softplus & APL   & Comb  & PAU   & PReLU & SLAF  & AReLU          \\ \hline\hline
    (5, 1)        &  82.50          & 81.88 & 83.13  & 70.00  & 83.75   & 25.63    & 71.25 & 75.63 & 43.13 & 88.13 & 84.38 & \textbf{92.50} \\ \hline
    (5, 5)        &  \textbf{94.30} & 69.37 & 93.12  & 78.00  & 93.50   & 22.12    & 63.00 & 88.25 & 40.12 & 91.75 & 77.00 & \textbf{94.30}
    \end{tabular}
    }}
    \label{tab:meta}\vspace{-8pt}
\end{table}

\section{Conclusion}
We have presented AReLU, a new learnable activation function formulated with element-wise sign-based attention mechanism. Networks implemented with AReLU can better mitigate the gradient vanishing issue and converge faster with small learning rates. This makes it especially useful in transfer learning where a pretrained model needs to be finetuned in the target domain with a small learning rate.
AReLU can significantly boost the performance of most mainstream network
architectures with only two extra learnable parameters per layer introduced.
In the future, we would like to investigate the application/extension of AReLU to more diverse tasks such as object detection, language translation and even structural feature learning with graph neural networks. 


\bibliographystyle{iclr2021_conference}
\bibliography{ref}

\begin{thebibliography}{45}
\providecommand{\natexlab}[1]{#1}
\providecommand{\url}[1]{\texttt{#1}}
\expandafter\ifx\csname urlstyle\endcsname\relax
  \providecommand{\doi}[1]{doi: #1}\else
  \providecommand{\doi}{doi: \begingroup \urlstyle{rm}\Url}\fi

\bibitem[Agostinelli et~al.(2014)Agostinelli, Hoffman, Sadowski, and
  Baldi]{agostinelli2014learning}
Forest Agostinelli, Matthew Hoffman, Peter Sadowski, and Pierre Baldi.
\newblock Learning activation functions to improve deep neural networks.
\newblock \emph{arXiv preprint arXiv:1412.6830}, 2014.

\bibitem[Ahmed et~al.(2017)Ahmed, Keskar, and Socher]{1711.02132}
Karim Ahmed, Nitish~Shirish Keskar, and Richard Socher.
\newblock Weighted transformer network for machine translation, 2017.

\bibitem[Bahdanau et~al.(2014)Bahdanau, Cho, and Bengio]{1409.0473}
Dzmitry Bahdanau, Kyunghyun Cho, and Yoshua Bengio.
\newblock Neural machine translation by jointly learning to align and
  translate, 2014.

\bibitem[Barron(2017)]{barron2017continuously}
Jonathan~T Barron.
\newblock Continuously differentiable exponential linear units.
\newblock \emph{arXiv preprint arXiv:1704.07483}, 2017.

\bibitem[Bochkovskiy et~al.(2020)Bochkovskiy, Wang, and
  Liao]{bochkovskiy2020yolov4}
Alexey Bochkovskiy, Chien-Yao Wang, and Hong-Yuan~Mark Liao.
\newblock Yolov4: Optimal speed and accuracy of object detection.
\newblock \emph{arXiv preprint arXiv:2004.10934}, 2020.

\bibitem[Buda et~al.(2019)Buda, Saha, and Mazurowski]{buda2019association}
Mateusz Buda, Ashirbani Saha, and Maciej~A Mazurowski.
\newblock Association of genomic subtypes of lower-grade gliomas with shape
  features automatically extracted by a deep learning algorithm.
\newblock \emph{Computers in biology and medicine}, 109:\penalty0 218--225,
  2019.

\bibitem[Chen et~al.(2017)Chen, Zhang, Xiao, Nie, Shao, Liu, and
  Chua]{chen2017sca}
Long Chen, Hanwang Zhang, Jun Xiao, Liqiang Nie, Jian Shao, Wei Liu, and
  Tat-Seng Chua.
\newblock {SCA-CNN}: Spatial and channel-wise attention in convolutional
  networks for image captioning.
\newblock In \emph{Proceedings of the IEEE conference on computer vision and
  pattern recognition}, pp.\  5659--5667, 2017.

\bibitem[Chorowski et~al.(2015)Chorowski, Bahdanau, Serdyuk, Cho, and
  Bengio]{chorowski2015attention}
Jan~K Chorowski, Dzmitry Bahdanau, Dmitriy Serdyuk, Kyunghyun Cho, and Yoshua
  Bengio.
\newblock Attention-based models for speech recognition.
\newblock In \emph{Advances in neural information processing systems}, pp.\
  577--585, 2015.

\bibitem[Clevert et~al.(2015)Clevert, Unterthiner, and
  Hochreiter]{clevert2015fast}
Djork-Arn{\'e} Clevert, Thomas Unterthiner, and Sepp Hochreiter.
\newblock Fast and accurate deep network learning by exponential linear units
  (elus).
\newblock \emph{arXiv preprint arXiv:1511.07289}, 2015.

\bibitem[Finn et~al.(2017)Finn, Abbeel, and Levine]{finn2017model}
Chelsea Finn, Pieter Abbeel, and Sergey Levine.
\newblock Model-agnostic meta-learning for fast adaptation of deep networks.
\newblock \emph{arXiv preprint arXiv:1703.03400}, 2017.

\bibitem[Glorot et~al.(2011)Glorot, Bordes, and Bengio]{glorot2011deep}
Xavier Glorot, Antoine Bordes, and Yoshua Bengio.
\newblock Deep sparse rectifier neural networks.
\newblock In \emph{Proceedings of the fourteenth international conference on
  artificial intelligence and statistics}, pp.\  315--323, 2011.

\bibitem[Goodfellow et~al.(2013)Goodfellow, Warde-Farley, Mirza, Courville, and
  Bengio]{goodfellow2013maxout}
Ian~J Goodfellow, David Warde-Farley, Mehdi Mirza, Aaron Courville, and Yoshua
  Bengio.
\newblock Maxout networks.
\newblock \emph{arXiv preprint arXiv:1302.4389}, 2013.

\bibitem[Goyal et~al.(2019)Goyal, Goyal, and Lall]{1906.09529}
Mohit Goyal, Rajan Goyal, and Brejesh Lall.
\newblock Learning activation functions: A new paradigm for understanding
  neural networks, 2019.

\bibitem[He et~al.(2015)He, Zhang, Ren, and Sun]{he2015delving}
Kaiming He, Xiangyu Zhang, Shaoqing Ren, and Jian Sun.
\newblock Delving deep into rectifiers: Surpassing human-level performance on
  imagenet classification.
\newblock In \emph{Proceedings of the IEEE international conference on computer
  vision}, pp.\  1026--1034, 2015.

\bibitem[He et~al.(2016)He, Zhang, Ren, and Sun]{he2016deep}
Kaiming He, Xiangyu Zhang, Shaoqing Ren, and Jian Sun.
\newblock Deep residual learning for image recognition.
\newblock In \emph{Proceedings of the IEEE conference on computer vision and
  pattern recognition}, pp.\  770--778, 2016.

\bibitem[Hendrycks \& Gimpel(2016{\natexlab{a}})Hendrycks and
  Gimpel]{1606.08415}
Dan Hendrycks and Kevin Gimpel.
\newblock Gaussian error linear units (gelus), 2016{\natexlab{a}}.

\bibitem[Hendrycks \& Gimpel(2016{\natexlab{b}})Hendrycks and
  Gimpel]{hendrycks2016gaussian}
Dan Hendrycks and Kevin Gimpel.
\newblock Gaussian error linear units (gelus).
\newblock \emph{arXiv preprint arXiv:1606.08415}, 2016{\natexlab{b}}.

\bibitem[Howard et~al.(2017)Howard, Zhu, Chen, Kalenichenko, Wang, Weyand,
  Andreetto, and Adam]{howard2017mobilenets}
Andrew~G Howard, Menglong Zhu, Bo~Chen, Dmitry Kalenichenko, Weijun Wang,
  Tobias Weyand, Marco Andreetto, and Hartwig Adam.
\newblock Mobilenets: Efficient convolutional neural networks for mobile vision
  applications.
\newblock \emph{arXiv preprint arXiv:1704.04861}, 2017.

\bibitem[Iandola et~al.(2016)Iandola, Han, Moskewicz, Ashraf, Dally, and
  Keutzer]{iandola2016squeezenet}
Forrest~N Iandola, Song Han, Matthew~W Moskewicz, Khalid Ashraf, William~J
  Dally, and Kurt Keutzer.
\newblock Squeezenet: Alexnet-level accuracy with 50x fewer parameters and< 0.5
  mb model size.
\newblock \emph{arXiv preprint arXiv:1602.07360}, 2016.

\bibitem[Kim et~al.(2019)Kim, Kim, Kang, and Lee]{1907.10830}
Junho Kim, Minjae Kim, Hyeonwoo Kang, and Kwanghee Lee.
\newblock U-gat-it: Unsupervised generative attentional networks with adaptive
  layer-instance normalization for image-to-image translation, 2019.

\bibitem[Klambauer et~al.(2017)Klambauer, Unterthiner, Mayr, and
  Hochreiter]{klambauer2017self}
G{\"u}nter Klambauer, Thomas Unterthiner, Andreas Mayr, and Sepp Hochreiter.
\newblock Self-normalizing neural networks.
\newblock In \emph{Advances in neural information processing systems}, pp.\
  971--980, 2017.

\bibitem[Krizhevsky \& Hinton(2010)Krizhevsky and
  Hinton]{krizhevsky2010convolutional}
Alex Krizhevsky and Geoff Hinton.
\newblock Convolutional deep belief networks on cifar-10.
\newblock \emph{Unpublished manuscript}, 40\penalty0 (7):\penalty0 1--9, 2010.

\bibitem[Krizhevsky et~al.(2009)Krizhevsky, Hinton,
  et~al.]{krizhevsky2009learning}
Alex Krizhevsky, Geoffrey Hinton, et~al.
\newblock Learning multiple layers of features from tiny images.
\newblock 2009.

\bibitem[LeCun et~al.(1998)LeCun, Bottou, Bengio, and
  Haffner]{lecun1998gradient}
Yann LeCun, L{\'e}on Bottou, Yoshua Bengio, and Patrick Haffner.
\newblock Gradient-based learning applied to document recognition.
\newblock \emph{Proceedings of the IEEE}, 86\penalty0 (11):\penalty0
  2278--2324, 1998.

\bibitem[Luong et~al.(2015)Luong, Pham, and Manning]{luong2015effective}
Minh-Thang Luong, Hieu Pham, and Christopher~D Manning.
\newblock Effective approaches to attention-based neural machine translation.
\newblock \emph{arXiv preprint arXiv:1508.04025}, 2015.

\bibitem[Ma et~al.(2018)Ma, Zhang, Zheng, and Sun]{ma2018shufflenet}
Ningning Ma, Xiangyu Zhang, Hai-Tao Zheng, and Jian Sun.
\newblock Shufflenet v2: Practical guidelines for efficient cnn architecture
  design.
\newblock In \emph{Proceedings of the European Conference on Computer Vision
  (ECCV)}, pp.\  116--131, 2018.

\bibitem[Maas et~al.(2013)Maas, Hannun, and Ng]{maas2013rectifier}
Andrew~L Maas, Awni~Y Hannun, and Andrew~Y Ng.
\newblock Rectifier nonlinearities improve neural network acoustic models.
\newblock In \emph{Proc. icml}, volume~30, pp.\ ~3, 2013.

\bibitem[Manessi \& Rozza(2018)Manessi and Rozza]{manessi2018learning}
Franco Manessi and Alessandro Rozza.
\newblock Learning combinations of activation functions.
\newblock In \emph{2018 24th International Conference on Pattern Recognition
  (ICPR)}, pp.\  61--66. IEEE, 2018.

\bibitem[Molina et~al.(2019)Molina, Schramowski, and Kersting]{molina2019pad}
Alejandro Molina, Patrick Schramowski, and Kristian Kersting.
\newblock Pad$\backslash$'e activation units: End-to-end learning of flexible
  activation functions in deep networks.
\newblock \emph{arXiv preprint arXiv:1907.06732}, 2019.

\bibitem[Nair \& Hinton(2010)Nair and Hinton]{nair2010rectified}
Vinod Nair and Geoffrey~E Hinton.
\newblock Rectified linear units improve restricted boltzmann machines.
\newblock In \emph{Proceedings of the 27th international conference on machine
  learning (ICML-10)}, pp.\  807--814, 2010.

\bibitem[Pereyra et~al.(2017)Pereyra, Tucker, Chorowski, Kaiser, and
  Hinton]{pereyra2017regularizing}
Gabriel Pereyra, George Tucker, Jan Chorowski, {\L}ukasz Kaiser, and Geoffrey
  Hinton.
\newblock Regularizing neural networks by penalizing confident output
  distributions.
\newblock \emph{arXiv preprint arXiv:1701.06548}, 2017.

\bibitem[Ramachandran et~al.(2017)Ramachandran, Zoph, and
  Le]{ramachandran2017searching}
Prajit Ramachandran, Barret Zoph, and Quoc~V Le.
\newblock Searching for activation functions.
\newblock \emph{arXiv preprint arXiv:1710.05941}, 2017.

\bibitem[Ronneberger et~al.(2015)Ronneberger, Fischer, and
  Brox]{ronneberger2015u}
Olaf Ronneberger, Philipp Fischer, and Thomas Brox.
\newblock U-net: Convolutional networks for biomedical image segmentation.
\newblock In \emph{International Conference on Medical image computing and
  computer-assisted intervention}, pp.\  234--241. Springer, 2015.

\bibitem[Selvaraju et~al.(2017)Selvaraju, Cogswell, Das, Vedantam, Parikh, and
  Batra]{selvaraju2017grad}
Ramprasaath~R Selvaraju, Michael Cogswell, Abhishek Das, Ramakrishna Vedantam,
  Devi Parikh, and Dhruv Batra.
\newblock Grad-cam: Visual explanations from deep networks via gradient-based
  localization.
\newblock In \emph{Proceedings of the IEEE international conference on computer
  vision}, pp.\  618--626, 2017.

\bibitem[Shin et~al.(2017)Shin, Lee, Kim, and Kim]{shin2017continual}
Hanul Shin, Jung~Kwon Lee, Jaehong Kim, and Jiwon Kim.
\newblock Continual learning with deep generative replay.
\newblock In \emph{Advances in Neural Information Processing Systems}, pp.\
  2990--2999, 2017.

\bibitem[Simonyan \& Zisserman(2014)Simonyan and Zisserman]{simonyan2014very}
Karen Simonyan and Andrew Zisserman.
\newblock Very deep convolutional networks for large-scale image recognition.
\newblock \emph{arXiv preprint arXiv:1409.1556}, 2014.

\bibitem[Tang et~al.(2019{\natexlab{a}})Tang, Liu, Xu, Torr, and
  Sebe]{tang2019attentiongan}
Hao Tang, Hong Liu, Dan Xu, Philip~HS Torr, and Nicu Sebe.
\newblock Attentiongan: Unpaired image-to-image translation using
  attention-guided generative adversarial networks.
\newblock \emph{arXiv preprint arXiv:1911.11897}, 2019{\natexlab{a}}.

\bibitem[Tang et~al.(2019{\natexlab{b}})Tang, Xu, Sebe, Wang, Corso, and
  Yan]{tang2019multichannel}
Hao Tang, Dan Xu, Nicu Sebe, Yanzhi Wang, Jason~J. Corso, and Yan Yan.
\newblock Multi-channel attention selection gan with cascaded semantic guidance
  for cross-view image translation.
\newblock In \emph{CVPR}, 2019{\natexlab{b}}.

\bibitem[Tang et~al.(2019{\natexlab{c}})Tang, Xu, Sebe, and
  Yan]{tang2019attention}
Hao Tang, Dan Xu, Nicu Sebe, and Yan Yan.
\newblock Attention-guided generative adversarial networks for unsupervised
  image-to-image translation.
\newblock In \emph{International Joint Conference on Neural Networks (IJCNN)},
  2019{\natexlab{c}}.

\bibitem[Tang et~al.(2020)Tang, Xu, Yan, Corso, Torr, and Sebe]{tang2020multi}
Hao Tang, Dan Xu, Yan Yan, Jason~J Corso, Philip~HS Torr, and Nicu Sebe.
\newblock Multi-channel attention selection gans for guided image-to-image
  translation.
\newblock \emph{arXiv preprint arXiv:2002.01048}, 2020.

\bibitem[Vaswani et~al.(2017)Vaswani, Shazeer, Parmar, Uszkoreit, Jones, Gomez,
  Kaiser, and Polosukhin]{vaswani2017attention}
Ashish Vaswani, Noam Shazeer, Niki Parmar, Jakob Uszkoreit, Llion Jones,
  Aidan~N Gomez, {\L}ukasz Kaiser, and Illia Polosukhin.
\newblock Attention is all you need.
\newblock In \emph{Advances in neural information processing systems}, pp.\
  5998--6008, 2017.

\bibitem[Xie et~al.(2017)Xie, Girshick, Doll{\'a}r, Tu, and
  He]{xie2017aggregated}
Saining Xie, Ross Girshick, Piotr Doll{\'a}r, Zhuowen Tu, and Kaiming He.
\newblock Aggregated residual transformations for deep neural networks.
\newblock In \emph{Proceedings of the IEEE conference on computer vision and
  pattern recognition}, pp.\  1492--1500, 2017.

\bibitem[Xu et~al.(2015{\natexlab{a}})Xu, Wang, Chen, and Li]{xu2015empirical}
Bing Xu, Naiyan Wang, Tianqi Chen, and Mu~Li.
\newblock Empirical evaluation of rectified activations in convolutional
  network.
\newblock \emph{arXiv preprint arXiv:1505.00853}, 2015{\natexlab{a}}.

\bibitem[Xu et~al.(2015{\natexlab{b}})Xu, Ba, Kiros, Cho, Courville,
  Salakhudinov, Zemel, and Bengio]{xu2015show}
Kelvin Xu, Jimmy Ba, Ryan Kiros, Kyunghyun Cho, Aaron Courville, Ruslan
  Salakhudinov, Rich Zemel, and Yoshua Bengio.
\newblock Show, attend and tell: Neural image caption generation with visual
  attention.
\newblock In \emph{International conference on machine learning}, pp.\
  2048--2057, 2015{\natexlab{b}}.

\bibitem[Xu et~al.(2015{\natexlab{c}})Xu, Ba, Kiros, Cho, Courville,
  Salakhutdinov, Zemel, and Bengio]{sat}
Kelvin Xu, Jimmy Ba, Ryan Kiros, Kyunghyun Cho, Aaron Courville, Ruslan
  Salakhutdinov, Richard Zemel, and Y.~Bengio.
\newblock Show, attend and tell: Neural image caption generation with visual
  attention.
\newblock 02 2015{\natexlab{c}}.

\end{thebibliography}

\clearpage

\begin{appendices}

\section{Details of MNIST-Conv}
\begin{figure}[h]
    \centering
    \includegraphics[width=\textwidth]{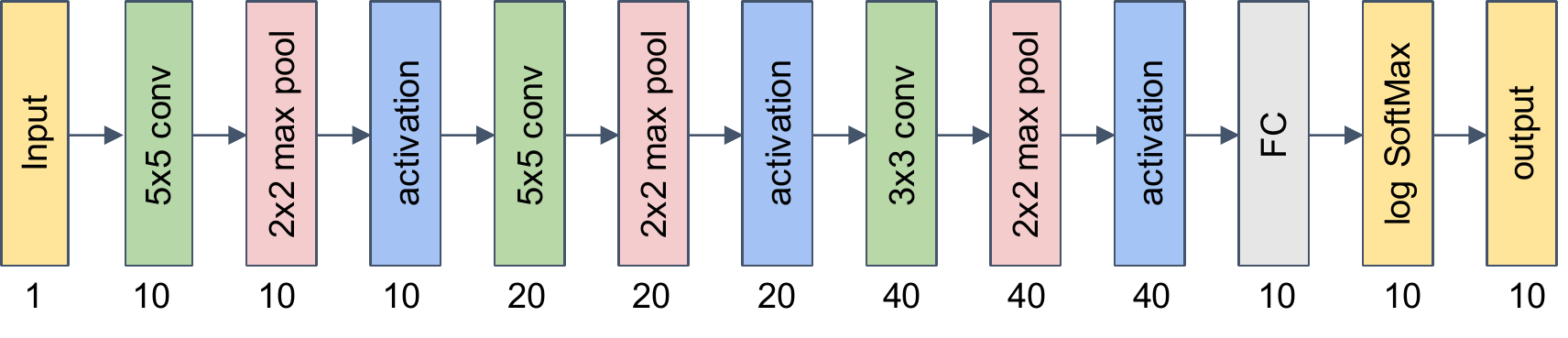}
	\caption{The network architecture of MNIST-Conv. The number under the block indicates that output channels of current layer.}
    \label{fig:conv_mnist}\vspace{-8pt}
\end{figure}

MNIST-Conv is a VGG-like network~\citep{simonyan2014very} but with fewer layers, as shown in Figure~\ref{fig:conv_mnist}.
The activation layers will be placed with specified activation functions while experiments.

\section{More results on MNIST}
In Table~\ref{tab:mnist-1st-best}, best testing accuracy on MNIST for five trainings of MNIST-Conv after the \emph{first epoch} with different optimizers and learning rates are reported.
In Table~\ref{tab:mnist-20-mean}, mean testing accuracy of five-time training of MNIST-Conv trained for $20$ epochs with different learning rates on MNIST are reported.
in Table~\ref{tab:mnist-20-best}, best testing accuracy of five-time training of MNIST-Conv trained for $20$ epochs with different learning rates on MNIST are reported.
We compare AReLU with $13$ non-learnable and $5$ learnable activation functions. The number of parameters per activation unit are listed beside the name of the learnable activation functions.
The best numbers are shown in bold text with blue color for non-learnable methods and red for learnable ones.
At the bottom of the table, we report the improvement of AReLU over the best among other non-learnable and learnable methods, in blue and red respectively.

\begin{figure}
    \centering
    \includegraphics[width=\textwidth]{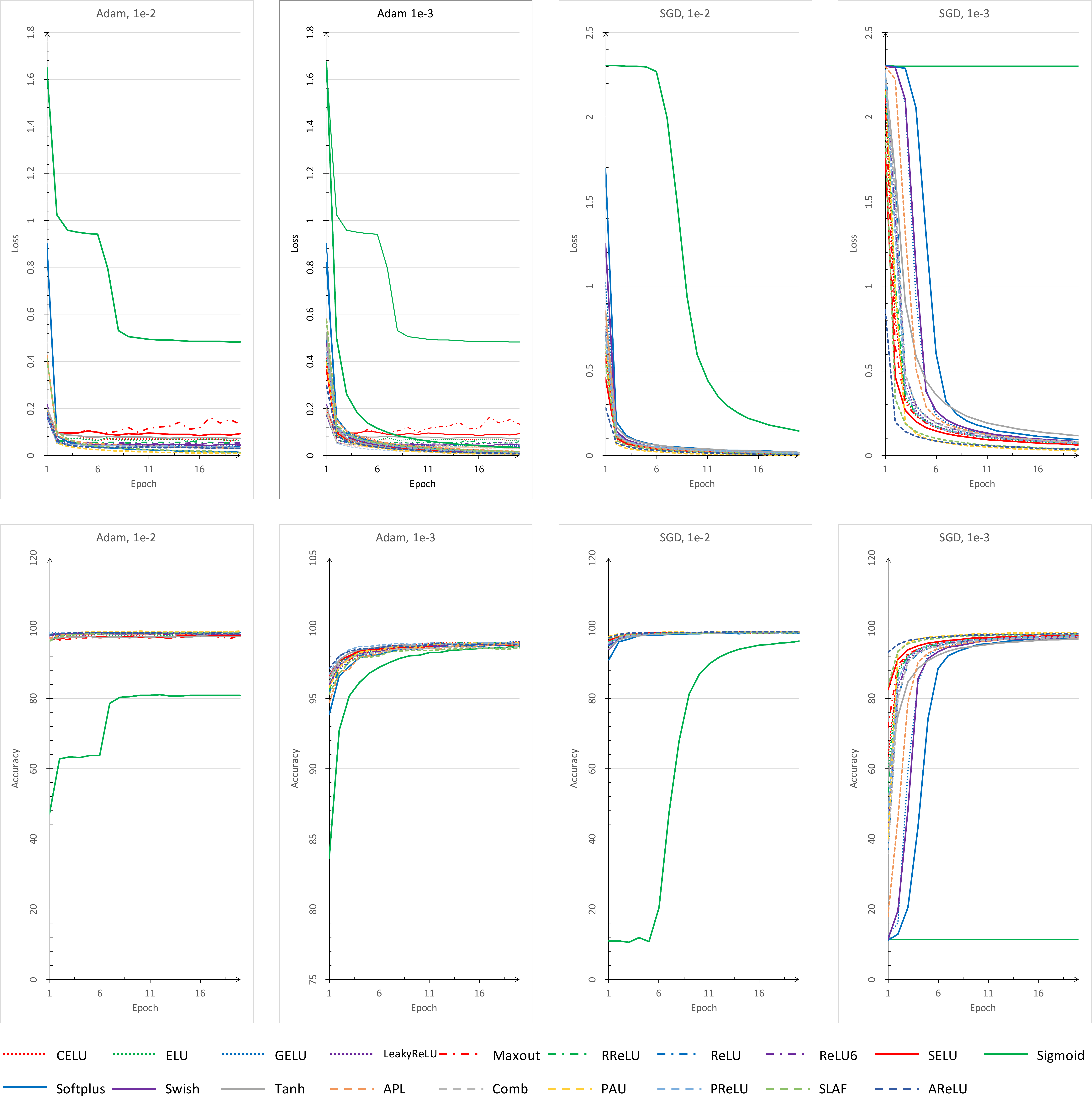}
	\caption{The plots of mean training loss and testing accuracy (\%) on MNIST for five-time trainings of MNIST-Conv over increasing training epochs with different optimizers and learning rates.}
    \label{fig:mnistplot_large}\vspace{-8pt}
\end{figure}

\begin{figure}
    \centering
    \includegraphics[width=\textwidth]{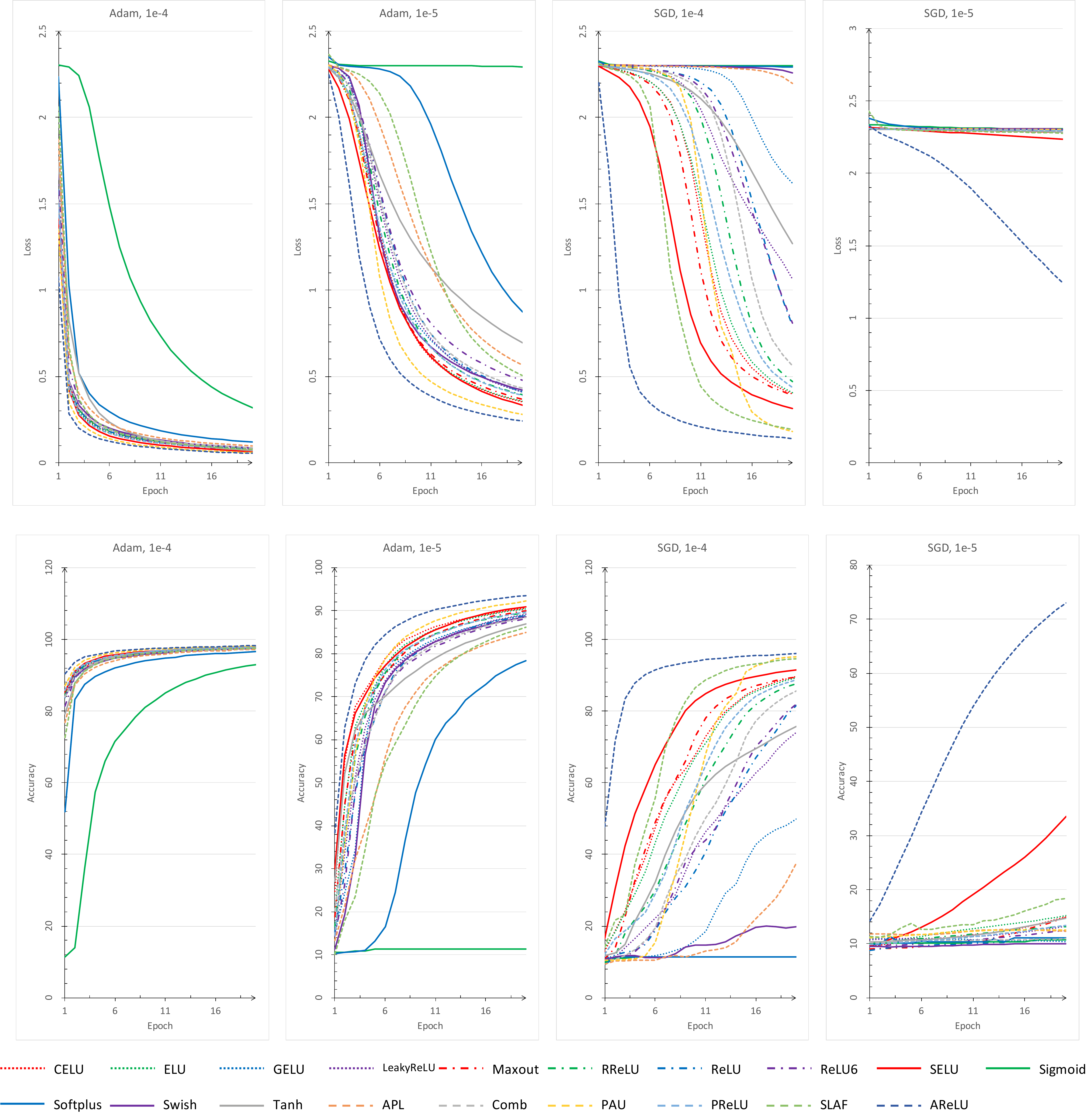}
	\caption{The plots of mean training loss and testing accuracy (\%) on MNIST for five-time trainings of MNIST-Conv over increasing training epochs with different optimizers and learning rates.}
    \label{fig:mnistplot_small}\vspace{-8pt}
\end{figure}

At the meantime, we also plot the mean training loss and testing accuracy of five runs with different optimizers and learning rates in Figure~\ref{fig:mnistplot_large} and Figure~\ref{fig:mnistplot_small}.

\begin{table}
	\centering
	\caption{Best testing accuracy (\%) on MNIST for five trainings of MNIST-Conv after the \emph{first epoch} with different optimizers and learning rates. We compare AReLU with $13$ non-learnable and $5$ learnable activation functions. The number of parameters per activation unit are listed beside the name of the learnable activation functions. The best numbers are shown in bold text with blue color for non-learnable methods and red for learnable ones. At the bottom of the table, we report the improvement of AReLU over the best among other non-learnable and learnable methods, in blue and red color respectively.}
	\label{tab:mnist-1st-best}
	\begin{tabular}{c|cc|cc|cc|cc}
		Learning Rate & \multicolumn{2}{c}{$1\times 10^{-2}$}                          & \multicolumn{2}{c|}{$1\times 10^{-3}$}                        & \multicolumn{2}{c|}{$1 \times 10 ^{-4}$}                       & \multicolumn{2}{c|}{$1\times 10^{-5}$}                      \\ \hline
		Optimizer                                & Adam            & SGD             & Adam            & SGD             & Adam            & SGD              & Adam             & SGD             \\ \hline \hline												
		CELU~\citeyearpar{barron2017continuously}       & 98.49           & 96.56           & 96.41           & 75.67           & 85.62           & 17.64            & 34.23            & 10.79           \\
		ELU~\citeyearpar{clevert2015fast}               & 98.36           & 96.64           & 96.34           & 65.66           & 86.77           & \bcblue{22.61}   & 28.91            & 11.36           \\
		GELU~\citeyearpar{hendrycks2016gaussian}        & \bcblue{98.68}  & 96.02           & 96.33           & 14.44           & 84.45           & 14.61            & 20.64            & \bcblue{13.30}  \\
		LReLU~\citeyearpar{maas2013rectifier}       & 98.29           & 95.93           & 96.19           & 44.76           & 84.86           & 13.20            & 20.55            & 11.49           \\
		Maxout~\citeyearpar{goodfellow2013maxout}       & 97.79           & 96.09           & 96.45           & 79.21           & 85.62           & 13.99            & 22.05            & 10.55           \\
		ReLU~\citeyearpar{nair2010rectified}            & 98.13           & 96.33           & 96.07           & 49.78           & 86.07           & 12.67            & 19.87            & 10.24           \\
		ReLU6~\citeyearpar{krizhevsky2010convolutional} & 98.18           & 96.05           & 96.55           & 56.07           & 83.42           & 13.13            & 17.42            & 10.27           \\
		RReLU~\citeyearpar{xu2015empirical}             & 98.52           & 96.30           & 95.98           & 61.78           & 86.97           & 10.36            & 20.61            & 11.35           \\
		SELU~\citeyearpar{klambauer2017self}            & 97.72           & \bcblue{96.88}  & \bcblue{97.01}  & \bcblue{83.85}  & \bcblue{87.53}  & 22.09            & \bcblue{37.98}   & 10.59           \\
		Sigmoid                                  & 97.62           & 11.35           & 85.29           & 11.35           & 11.47           & 11.35            & 11.35            & 10.28           \\
		Softplus~\citeyearpar{glorot2011deep}           & 97.80           & 93.83           & 94.58           & 11.35           & 75.05           & 11.35            & 11.35            & 10.32           \\
		Swish~\citeyearpar{ramachandran2017searching}   & 98.32           & 95.28           & 96.47           & 12.38           & 85.52           & 11.82            & 15.51            & 10.27           \\
		Tanh                                     & 97.32           & 94.40           & 96.84           & 69.29           & 81.50           & 16.32            & 29.92            & 11.35           \\ \hline \hline			
		APL~\citeyearpar{agostinelli2014learning} ($2$) & \bcred{98.48}   & 96.25           & 95.50           & 27.75           & 80.56           & 10.28            & 19.50            & 15.47           \\
		Comb~\citeyearpar{manessi2018learning} ($1$)    & 98.42           & 96.54           & 96.07           & 57.88           & 85.59           & 11.67            & 25.40            & 10.72           \\
		PAU~\citeyearpar{molina2019pad} ($10$)          & 98.42           & \bcred{97.94}   & 97.07           & 76.69           & 89.71           & 11.35            & 18.11            & 14.54           \\
		PReLU~\citeyearpar{he2015delving} ($1$)         & 98.52           & 96.10           & 96.33           & 61.72           & 87.24           & 15.86            & 18.31            & 11.40           \\
		SLAF~\citeyearpar{1906.09529} ($2$)             & 96.69           & 97.27           & 95.76           & 84.13           & 76.84           & 15.60            & 11.19            & 13.09           \\ \hline
		AReLU ($2$)                              & 98.46           & 97.60           & \bcred{97.29}   & \bcred{93.83}   & \bcred{90.91}   & \bcred{61.06}    & \bcred{48.06}    & \bcred{19.84}   \\
		\cblue{Improvement}                      & \cblue{$-0.22$} & \cblue{$+0.72$} & \cblue{$+0.28$} & \cblue{$+9.98$} & \cblue{$+3.38$} & \cblue{+$38.45$} & \cblue{$+10.08$} & \cblue{$+6.54$} \\
		\cred{Improvement}                       & \cred{$-0.02$}  & \cred{$-0.34$}  & \cred{$+0.22$}  & \cred{$+9.70$}  & \cred{$+1.20$}  & \cred{$+45.20$}  & \cred{$+22.66$}  & \cred{$+4.37$}  
	\end{tabular}
\end{table}
\begin{table}
	\centering
	\caption{Best testing accuracy (\%) of five-time training of MNIST-Conv trained for $20$ epochs with different learning rates on MNIST. We compare AReLU with $13$ non-learnable and $5$ learnable activation functions. The number of parameters per activation unit are listed beside the name of the learnable activation functions. The best numbers are shown in bold text with blue color for non-learnable methods and red for learnable ones. At the bottom of the table, we report the improvement of AReLU over the best among other non-learnable and learnable methods, in blue and red respectively.}
	\label{tab:mnist-20-best}
	\begin{tabular}{c|cc|cc|cc|cc}
		Learning Rate & \multicolumn{2}{c}{$1\times 10^{-2}$}                          & \multicolumn{2}{c|}{$1\times 10^{-3}$}                        & \multicolumn{2}{c|}{$1 \times 10 ^{-4}$}                       & \multicolumn{2}{c|}{$1\times 10^{-5}$}                      \\ \hline
		Optimizer                                & Adam            & SGD             & Adam            & SGD             & Adam            & SGD             & Adam            & SGD              \\ \hline \hline
		CELU~\citeyearpar{barron2017continuously}       & 98.67           & 99.05           & 99.14           & 97.98           & 97.88           & 90.89           & 91.33           & 17.39            \\
		ELU~\citeyearpar{clevert2015fast}               & 98.70           & 99.04           & 99.10           & 97.93           & 97.96           & 90.84           & 91.40           & 20.51            \\
		GELU~\citeyearpar{hendrycks2016gaussian}        & 99.03           & 98.99           & 99.14           & 97.65           & 98.04           & 85.60           & 90.19           & 13.30            \\
		LReLU~\citeyearpar{maas2013rectifier}       & 98.79           & 99.04           & 99.10           & 97.85           & 97.91           & 90.00           & 90.90           & 16.63            \\
		Maxout~\citeyearpar{goodfellow2013maxout}       & 98.30           & 98.86           & 98.82           & 97.98           & 97.69           & 89.98           & 91.04           & 23.90            \\
		ReLU~\citeyearpar{nair2010rectified}            & 98.80           & \bcblue{99.06}  & 99.17           & 97.64           & 97.85           & 86.53           & 89.98           & 13.95            \\
		ReLU6~\citeyearpar{krizhevsky2010convolutional} & 98.55           & 99.01           & 99.17           & 98.03           & \bcblue{98.14}  & 86.57           & 88.98           & 18.79            \\
		RReLU~\citeyearpar{xu2015empirical}             & 98.98           & 99.05           & \bcblue{99.19}  & 97.90           & 97.76           & 88.23           & 90.31           & 18.31            \\
		SELU~\citeyearpar{klambauer2017self}            & 98.53           & 98.96           & 98.91           & \bcblue{98.04}  & 98.06           & \bcblue{92.09}  & \bcblue{92.26}  & \bcblue{37.85}   \\
		Sigmoid                                  & 98.99           & 96.37           & 98.78           & 11.35           & 94.02           & 11.35           & 11.35           & 11.35            \\
		Softplus~\citeyearpar{glorot2011deep}           & \bcblue{99.07}  & 98.86           & 99.04           & 97.52           & 96.86           & 11.88           & 80.60           & 16.34            \\
		Swish~\citeyearpar{ramachandran2017searching}   & 98.83           & 98.85           & 99.09           & 97.65           & 97.80           & 36.30           & 89.21           & 10.29            \\
		Tanh                                     & 97.96           & 98.89           & 99.02           & 97.27           & 98.09           & 78.30           & 88.17           & 17.76            \\ \hline \hline									        
		APL~\citeyearpar{agostinelli2014learning} ($2$) & 98.80           & 99.02           & 99.00           & 97.73           & 97.35           & 49.37           & 87.24           & 16.14            \\
		Comb~\citeyearpar{manessi2018learning} ($1$)    & 99.03           & 99.10           & 99.16           & 97.71           & 97.90           & 86.52           & 89.52           & 12.48            \\
		PAU~\citeyearpar{molina2019pad} ($10$)          & \bcred{99.19}   & 99.07           & \bcred{99.18}   & \bcred{98.82}   & 98.28           & 95.75           & 92.98           & 15.82            \\
		PReLU~\citeyearpar{he2015delving} ($1$)         & 98.97           & \bcred{99.11}   & 99.11           & 97.93           & 98.07           & 90.34           & 91.31           & 17.82            \\
		SLAF~\citeyearpar{1906.09529} ($2$)             & 98.88           & 98.97           & 98.82           & 98.50           & 97.82           & 94.88           & 87.67           & 25.35            \\ \hline 
		AReLU ($2$)                              & 99.08           & 99.07           & 99.05           & 98.60           & \bcred{98.40}   & \bcred{96.32}   & \bcred{93.75}   & \bcred{85.45}    \\
		\cblue{Improvement}                      & \cblue{$+0.01$} & \cblue{$+0.01$} & \cblue{$-0.14$} & \cblue{$+0.56$} & \cblue{$+0.26$} & \cblue{+$4.23$} & \cblue{$+1.49$} & \cblue{$+47.60$} \\
		\cred{Improvement}                       & \cred{$-0.11$}  & \cred{$-0.04$}  & \cred{$-0.13$}  & \cred{$-0.22$}  & \cred{$+0.12$}  & \cred{$+0.57$}  & \cred{$+0.77$}  & \cred{$+60.10$}  
	\end{tabular}
\end{table}
\begin{table}
	\centering
	\caption{Mean testing accuracy (\%) of five-time training of MNIST-Conv trained for $20$ epochs with different learning rates on MNIST. We compare AReLU with $13$ non-learnable and $5$ learnable activation functions. The number of parameters per activation unit are listed beside the name of the learnable activation functions. The best numbers are shown in bold text with blue color for non-learnable methods and red for learnable ones. At the bottom of the table, we report the improvement of AReLU over the best among other non-learnable and learnable methods, in blue and red respectively.}
	\label{tab:mnist-20-mean}
	\begin{tabular}{l|rr|rr|rr|rr}
		Learning Rate & \multicolumn{2}{c|}{$1\times 10^{-2}$}  & \multicolumn{2}{c|}{$1\times 10^{-3}$}              & \multicolumn{2}{c|}{$1 \times 10 ^{-4}$}    & \multicolumn{2}{c}{$1\times 10^{-5}$}         \\ \hline
		Optimizer                                & Adam            & SGD             & Adam            & SGD             & Adam            & SGD             & Adam            & SGD              \\ \hline\hline
		CELU~\citeyearpar{barron2017continuously}       & 98.62           & 98.93           & 99.05           & 97.73           & 97.70           & 89.58           & 90.58           & 14.96            \\
		ELU~\citeyearpar{clevert2015fast}               & 98.55           & 98.94           & 99.02           & 97.82           & 97.70           & 89.24           & 90.46           & 15.41            \\
		GELU~\citeyearpar{1606.08415}                   & 98.85           & 98.93           & \bcblue{99.08}  & 97.51           & 97.67           & 51.19           & 88.94           & 10.94            \\
		LReLU~\citeyearpar{maas2013rectifier}       & 98.66           & 98.92           & 98.96           & 97.74           & 97.61           & 74.01           & 89.21           & 13.27            \\
		Maxout~\citeyearpar{goodfellow2013maxout}       & 98.23           & 98.78           & 98.76           & 97.67           & 97.46           & 89.52           & 90.04           & 14.85            \\
		ReLU~\citeyearpar{nair2010rectified}            & 98.72           & \bcblue{98.98}  & 99.05           & 97.57           & 97.58           & 81.63           & 88.88           & 11.03            \\
		ReLU6~\citeyearpar{krizhevsky2010convolutional} & 98.51           & 98.93           & 99.02           & 97.79           & 97.96           & 81.96           & 88.14           & 12.44            \\
		RReLU~\citeyearpar{xu2015empirical}             & 98.78           & 98.94           & 99.06           & 97.77           & 97.62           & 87.64           & 89.59           & 13.37            \\
		SELU~\citeyearpar{klambauer2017self}            & 98.34           & 98.91           & 98.84           & \bcblue{97.98}  & 97.88           & \bcblue{91.56}  & \bcblue{90.91}  & \bcblue{33.48}   \\
		Sigmoid                                  & 81.05           & 96.24           & 98.72           & 11.35           & 92.96           & 11.35           & 11.35           & 10.67            \\
		Softplus~\citeyearpar{glorot2011deep}           & \bcblue{98.95}  & 98.78           & 98.93           & 97.30           & 96.57           & 11.52           & 78.36           & 12.50            \\
		Swish~\citeyearpar{ramachandran2017searching}   & 98.77           & 98.80           & 99.02           & 97.44           & 97.51           & 23.77           & 88.53           & 10.05            \\
		Tanh                                     & 97.91           & 98.86           & 98.96           & 96.94           & \bcblue{97.97}  & 75.66           & 86.99           & 14.76            \\ \hline \hline
		APL~\citeyearpar{agostinelli2014learning} ($2$) & 98.72           & 98.92           & 98.94           & 97.56           & 97.22           & 37.67           & 84.95           & 13.52            \\
		Comb~\citeyearpar{manessi2018learning} ($1$)    & 98.88           & \bcred{99.01}   & 99.04           & 97.56           & 97.55           & 85.60           & 88.39           & 10.94            \\
		PAU~\citeyearpar{molina2019pad} ($10$)          & \bcred{99.17}   & \bcred{99.01}   & \bcred{99.07}   & \bcred{98.78}   & 98.15           & 95.21           & 92.22           & 12.86            \\
		PReLU~\citeyearpar{he2015delving} ($1$)         & 98.89           & 98.86           & 99.01           & 97.81           & 97.77           & 88.67           & 89.69           & 13.36            \\
		SLAF~\citeyearpar{1906.09529} ($2$)             & 98.80           & 98.86           & 98.67           & 98.37           & 97.60           & 94.61           & 86.10           & 18.91            \\ \hline
		AReLU ($2$)                              & 98.94           & \bcred{99.01}   & 98.97           & 98.46           & \bcred{98.22}   & \bcred{96.00}   & \bcred{93.48}   & \bcred{73.00}    \\
		\cblue{Improvement}                      & \cblue{$-0.01$} & \cblue{$+0.03$} & \cblue{$-0.11$} & \cblue{$+0.48$} & \cblue{$+0.25$} & \cblue{$+4.44$} & \cblue{$+2.57$} & \cblue{$+39.52$} \\
		\cred{Improvement}                       & \cred{$-0.23$}  & \cred{$+0.00$}  & \cred{$-0.10$}  & \cred{$-0.32$}  & \cred{$+0.07$}  & \cred{$+0.79$}  & \cred{$+1.26$}  & \cred{$+54.09$}  
	\end{tabular}
\end{table} 

\section{Performance on CIFAR10}
In this experiment, we compare all the activation functions with three widely used network architectures on the CIFAR10 dataset. The three networks are VGG-11~\citep{simonyan2014very}, ResNet-18~\cite{he2016deep}, and MobileNet~\citep{howard2017mobilenets}.
We use the same training configuration for all the three networks. The initial learning rate is set to $0.1$ and is multiplied by $0.1$ at the $50$-th and $75$-th epochs. We train for $100$ epochs with the batch size being $128$, the weight decay being $5\times10^{-4}$, and the Nesterov momentum being $0.9$.

For each network with each activation function, we train five times and report both the mean and the best test accuracy in Table~\ref{tab:cifar10}.
From the comparison, AReLU performs the best with VGG-11, and reasonably well with ResNet-18 and MobileNet, with only two learnable parameters.
AReLU outperforms the other learnable activation functions with comparable number of learnable parameters.
AReLU works comparably well against PAU~\citep{molina2019pad} although the latter contains five times learnable parameters.
In conclusion, AReLU adapts well to different network architectures.

\section{More result on CIFAR100}
For each network with each activation function, we run five times of training and report the mean accuracy in top-1 and top-5 classification results; see Table~\ref{tab:cifar100}.
\begin{table}[htbp]
	\centering\small
	\caption{Mean test accuracy in top-1 and top-5 results of five times training on CIFAR100.}
	\label{tab:cifar100}
	\begin{tabular}{ccccccc}
		\multicolumn{1}{c|}{Accuracy (\%)} & @top-1                & \multicolumn{1}{c|}{@top-5}          & @top-1                & \multicolumn{1}{c|}{@top-5}          & @top-1                & @top-5                \\ \hline
		\multicolumn{1}{c|}{Network}      & \multicolumn{2}{c|}{VGG-11~\citeyearpar{simonyan2014very}}                                  & \multicolumn{2}{c|}{VGG-13~\citeyearpar{simonyan2014very}}                                 & \multicolumn{2}{c}{SEResNet18~\citeyearpar{xie2017aggregated}}             \\ \hline
		\multicolumn{1}{c|}{ReLU}     & 68.03                & \multicolumn{1}{c|}{87.49}          & 71.92                & \multicolumn{1}{c|}{90.05}          & 76.80                & 93.16                \\
		\multicolumn{1}{c|}{AReLU}        & \textbf{68.04}       & \multicolumn{1}{c|}{\textbf{88.27}} & \textbf{71.93}       & \multicolumn{1}{c|}{\textbf{90.86}} & \textbf{76.94}       & \textbf{93.66}       \\
		\multicolumn{1}{c|}{Improvement}      & $+0.01$       & \multicolumn{1}{c|}{$+0.78$} & $+0.01$       & \multicolumn{1}{c|}{$+0.81$} & $+0.14$       & $+0.50$       \\ \hline
		\multicolumn{1}{c|}{Network}      & \multicolumn{2}{c|}{ResNet-18~\citeyearpar{he2016deep}}                               & \multicolumn{2}{c|}{ShuffleNet-v2~\citeyearpar{ma2018shufflenet}}                          & \multicolumn{2}{c}{SqueezeNet~\citeyearpar{iandola2016squeezenet}}             \\ \hline
		\multicolumn{1}{c|}{ReLU}     & 76.35                & \multicolumn{1}{c|}{93.18}          & 68.56                & \multicolumn{1}{c|}{90.92}          & 69.96                & 91.29                \\
		\multicolumn{1}{c|}{AReLU}        & \textbf{76.40}       & \multicolumn{1}{c|}{\textbf{93.39}} & \textbf{69.63}       & \multicolumn{1}{c|}{\textbf{91.37}} & \textbf{70.19}       & \textbf{91.41}       \\
		\multicolumn{1}{c|}{Improvement}      & $+0.05$       & \multicolumn{1}{c|}{$+0.20$} & $+1.07$       & \multicolumn{1}{c|}{$+0.45$} & $+0.23$       & $+0.12$       \\
		\multicolumn{1}{l}{}              & \multicolumn{1}{l}{} & \multicolumn{1}{l}{}                & \multicolumn{1}{l}{} & \multicolumn{1}{l}{}                & \multicolumn{1}{l}{} & \multicolumn{1}{l}{}
	\end{tabular}\vspace{-8pt}
\end{table} 

\begin{table}
	\centering\small
	\caption{Test accuracy of five times training on CIFAR10.}
	\label{tab:cifar10}
	\begin{tabular}{l|cccccc}
        Network     & \multicolumn{2}{c|}{VGG-11~\citeyearpar{simonyan2014very}}                  & \multicolumn{2}{c|}{ResNet-18~\citeyearpar{he2016deep}}    & \multicolumn{2}{c}{MobileNet~\citeyearpar{howard2017mobilenets}}  \\ \hline
		Accuracy (\%)    & best   & \multicolumn{1}{c|}{mean}    & best   & \multicolumn{1}{c|}{mean}                    & best   & \multicolumn{1}{c}{mean}       \\ \hline\hline
		CELU~\citeyearpar{barron2017continuously}             & 89.76                   & \multicolumn{1}{c|}{89.92}                   & 92.02                   & \multicolumn{1}{c|}{91.87}                   & 90.07                   & \multicolumn{1}{c}{89.91}                   \\
		ELU~\citeyearpar{clevert2015fast}                     & 90.10                   & \multicolumn{1}{c|}{89.92}                   & 91.85                   & \multicolumn{1}{c|}{91.70}                   & 90.09                   & \multicolumn{1}{c}{89.80}                   \\
		GELU~\citeyearpar{hendrycks2016gaussian}                         & 91.54                   & \multicolumn{1}{c|}{91.29}                   & 94.29                   & \multicolumn{1}{c|}{94.18}                   & \bcblue{92.56} & \multicolumn{1}{c}{\bcblue{92.74}}          \\
		LReLU~\citeyearpar{maas2013rectifier}             & \bcblue{91.89} & \multicolumn{1}{c|}{\bcblue{91.75}}          & \bcblue{94.78}          & \multicolumn{1}{c|}{\bcblue{94.65}}          & 90.75                   & \multicolumn{1}{c}{90.65}                   \\
		Maxout~\citeyearpar{goodfellow2013maxout}             & 88.42                   & \multicolumn{1}{c|}{88.22}                   & $-$                      & \multicolumn{1}{c|}{$-$}                       & $-$                       & \multicolumn{1}{c}{$-$}                       \\
		RReLU~\citeyearpar{xu2015empirical}                   & 91.52                   & \multicolumn{1}{c|}{91.38}                   & 94.10                   & \multicolumn{1}{c|}{94.05}                   & 92.48                   & \multicolumn{1}{c}{92.20}                   \\
		ReLU~\citeyearpar{nair2010rectified}                  & 91.70                   & \multicolumn{1}{c|}{91.57}                   & 94.36                   & \multicolumn{1}{c|}{94.32}                   & 90.67                   & \multicolumn{1}{c}{90.52}                   \\
		ReLU6~\citeyearpar{krizhevsky2010convolutional}       & 91.73                   & \multicolumn{1}{c|}{91.48}                   & 94.74                   & \multicolumn{1}{c|}{94.59} & 90.69                   & \multicolumn{1}{c}{90.60}                   \\
		SELU~\citeyearpar{klambauer2017self}                  & 89.48                   & \multicolumn{1}{c|}{89.18}                   & 91.61                   & \multicolumn{1}{c|}{91.38}                   & 88.61                   & \multicolumn{1}{c}{88.56}                   \\
		Sigmoid                                       & $-$                       & \multicolumn{1}{c|}{$-$}                       & 81.11                   & \multicolumn{1}{c|}{80.49}                   & 79.37                   & \multicolumn{1}{c}{77.78}                   \\
		Softplus\citeyearpar{glorot2011deep}                 & 86.94                   & \multicolumn{1}{c|}{86.51}                   & 88.97                   & \multicolumn{1}{c|}{88.56}                   & 87.45                   & \multicolumn{1}{c}{87.10}                   \\
		Swish~\citeyearpar{ramachandran2017searching}                        & 90.84                   & \multicolumn{1}{c|}{90.75}                   & 93.67                   & \multicolumn{1}{c|}{93.57}                   & 91.94                   & \multicolumn{1}{c}{91.74}                   \\
		Tanh                                          & 90.22                   & \multicolumn{1}{c|}{89.99}                   & 91.61                   & \multicolumn{1}{c|}{91.50}                   & 88.74                   & \multicolumn{1}{c}{88.54}                   \\ \hline\hline
		APL~\citeyearpar{agostinelli2014learning} ($2$)             & 91.65                   & \multicolumn{1}{c|}{91.03}                   & 94.60                   & \multicolumn{1}{c|}{93.79}                   & 90.88                   & \multicolumn{1}{c}{90.07}                   \\
		Comb~\citeyearpar{manessi2018learning} ($1$)             & 90.90                   & \multicolumn{1}{c|}{63.32}                   & 93.82                   & \multicolumn{1}{c|}{93.28}                   & 92.04                   & \multicolumn{1}{c}{91.51}                   \\
		PAU~\citeyearpar{molina2019pad} ($10$)                      & 91.76                   & \multicolumn{1}{c|}{90.94}                   & \bcred{94.78}          & \multicolumn{1}{c|}{\bcred{94.52}}                   & \bcred{92.71}          & \multicolumn{1}{c}{\bcred{92.12}} \\
		PReLU~\citeyearpar{he2015delving} ($1$)                     & 91.13                   & \multicolumn{1}{c|}{90.11}                   & 93.82                   & \multicolumn{1}{c|}{93.61}                   & 91.71                   & \multicolumn{1}{c}{91.16}                   \\
		SLAF~\citeyearpar{1906.09529} ($2$)                         & $-$                       & \multicolumn{1}{c|}{$-$}                       & $-$                       & \multicolumn{1}{c|}{$-$}                       & $-$                       & \multicolumn{1}{c}{$-$}                       \\ \hline
		AReLU ($2$)                                         & \bcred{91.90}          & \multicolumn{1}{c|}{\bcred{91.59}} & \bcred{94.78} & \multicolumn{1}{c|}{94.43}                   & 91.68                   & \multicolumn{1}{c}{91.56}                   \\
		\cblue{Improvement}      & \cblue{$+0.01$}    & \multicolumn{1}{c|}{\cblue{$-0.16$}}   & \cblue{$+0.00$}          & \multicolumn{1}{c|}{\cblue{$-0.22$}}          & \cblue{$-0.88$}          & \multicolumn{1}{c}{\cblue{$-1.18$}} \\
\cred{Improvement}       & \cred{$+0.20$}    & \multicolumn{1}{c|}{\cred{$+0.56$}}    & \cred{$+0.00$}    & \multicolumn{1}{c|}{\cred{$-0.09$}}          & \cred{$-1.03$}          & \multicolumn{1}{c}{\cred{$-0.56$}}
	\end{tabular}\vspace{-8pt}
\end{table} 

\section{Performance in Image Segmentation}
\begin{figure}[h]
	\centering
	\includegraphics[width=\textwidth]{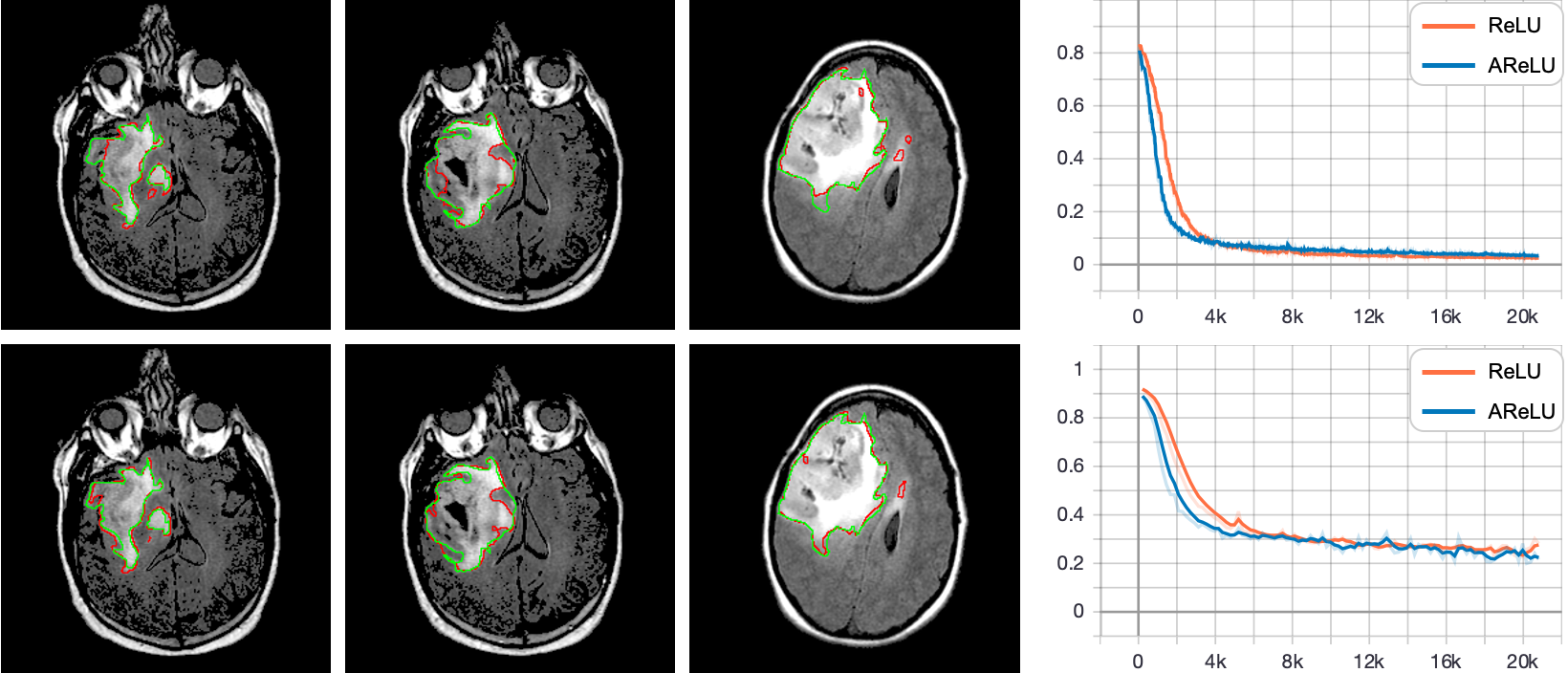}
	\caption{Left: Segmentation results of UNet with ReLU (top row) and AReLU (bottom). Prediction is depicted in red and ground-truth in green. Right: Training and validation loss over iterations.}
	\label{fig:seg}
\end{figure}

We test AReLU with UNet~\citep{ronneberger2015u} for brain image segmentation.
We use the Kaggle Brain MRI Segmentation dataset\footnote{https://www.kaggle.com/mateuszbuda/lgg-mri-segmentation} and follow the implementation details in~\citep{buda2019association}.
The dataset contains MRI from the TCIA LGG collection \footnote{https://wiki.cancerimagingarchive.net/display/Public/TCGA-LGG} with expert-approved segmentation masks.
Figure~\ref{fig:seg} shows that AReLU leads to a faster training than ReLU and achieves a better segmentation accuracy ($91.14\%$ vs. $90.77\%$ for the DSC metric~\citep{ronneberger2015u}).


\end{appendices}

\end{document}